%% file: neurips_2024.tex
\documentclass{article}

\usepackage[nonatbib,preprint]{neurips_2024}

\input{preamble.tex}

\usepackage{threeparttable}

\title{FedSheafHN: Personalized Federated Learning on Graph-structured Data}

\author{Wenfei Liang\textsuperscript{\rm 1}$^{\dagger}$$^{\ast}$, 
        Yanan Zhao\textsuperscript{\rm 1}$^{\dagger}$, 
        Rui She\textsuperscript{\rm 1}$^{\ast}$, 
        Yiming Li\textsuperscript{\rm 1}, 
        Wee Peng Tay\textsuperscript{\rm 1}\\
        \textsuperscript{\rm 1}Nanyang Technological University\\
        $^{\dagger}$Contributed equally \\
$^{\ast}$Corresponding author: \texttt{wenfei001@e.ntu.edu.sg}\\
}

\begin{document}

\maketitle

\begin{abstract} 
Personalized subgraph Federated Learning (FL) is a task that customizes Graph Neural Networks (GNNs) to individual client needs, accommodating diverse data distributions. 
However, applying hypernetworks in FL, while aiming to facilitate model personalization, often encounters challenges due to inadequate representation of client-specific characteristics. To overcome these limitations, we propose a model called \textit{FedSheafHN}, using enhanced collaboration graph embedding and efficient personalized model parameter generation. Specifically, our model embeds each client's local subgraph into a server-constructed collaboration graph. We utilize sheaf diffusion in the collaboration graph to learn client representations. Our model improves the integration and interpretation of complex client characteristics. Furthermore, our model ensures the generation of personalized models through advanced hypernetworks optimized for parallel operations across clients. 
Empirical evaluations demonstrate that FedSheafHN outperforms existing methods in most scenarios, in terms of client model performance on various graph-structured datasets. It also has fast model convergence and effective new clients generalization. 
\blue{This paper was submitted to ICML 2024 in Feb 2024. You can find a record of the submission here: }\href{https://github.com/CarrieWFF/ICML-2024-submission-recording/blob/main/Screenshot%20of%20FedSheafHN%20submission%20to%20ICML%202024.png}{\blue{Recording of FedSheafHN submission to ICML 2024}}.
The code is available at \href{https://github.com/CarrieWFF/FedSheafHN}{FedSheafHN\_srccode}.
\end{abstract}
\vspace{-3mm}
%#############################################################################
\section{Introduction}
\label{introduction}
\vspace{-2mm}
Many graph neural networks (GNNs) \cite{hamilton2020graph} focus on a single graph, storing nodes and edges from diverse sources in a central server. In practical scenarios, privacy or storage constraints lead to individual users or institutions maintaining private graphs. Collaborative training of GNNs across distributed graphs can be achieved through Federated Learning (FL) \cite{zhang2021subgraph,wu2021fedgnn,pillutla2022federated}, where each participant trains a local GNN \cite{wu2020comprehensive,bodnar2022neural,JiLeeMen23,KanZhaSon23}, and a central server aggregates their updated weights. In practice, each client may have unique subgraphs, and client data distributions can vary significantly. Some clients might even tackle distinct tasks. Personalized Federated Learning (PFL) \cite{smith2017federated} seeks to address this by allowing each client to use a personalized model instead of a shared one. 

The main challenge in PFL is to balance joint training benefits with maintaining unique models for each client. Many approaches integrate global model training with either client-specific fine-tuning or customization. However, this approach faces several challenges, such as inadequately addressing data heterogeneity among clients, not permitting heterogeneity in client models, and failing to quickly and effectively generalize to new clients. Hypernetworks (HNs) present a potential solution to these challenges. HNs are models designed to generate parameters for other neural networks, using a descriptive vector for that network as input \cite{ha2016hypernetworks}. Many FL frameworks employing HNs depend on basic vectors or partial model parameters \cite{shamsian2021personalized,xu2023heterogeneous}, which might not accurately reflect the unique characteristics of each client. However, the effectiveness of HNs hinges on a precise and detailed representation of each client’s data, highlighting the need for accurate descriptions to enable the generation of customized models for each client.

\begin{figure*}[t] 
\centering
\includegraphics[width=\textwidth]{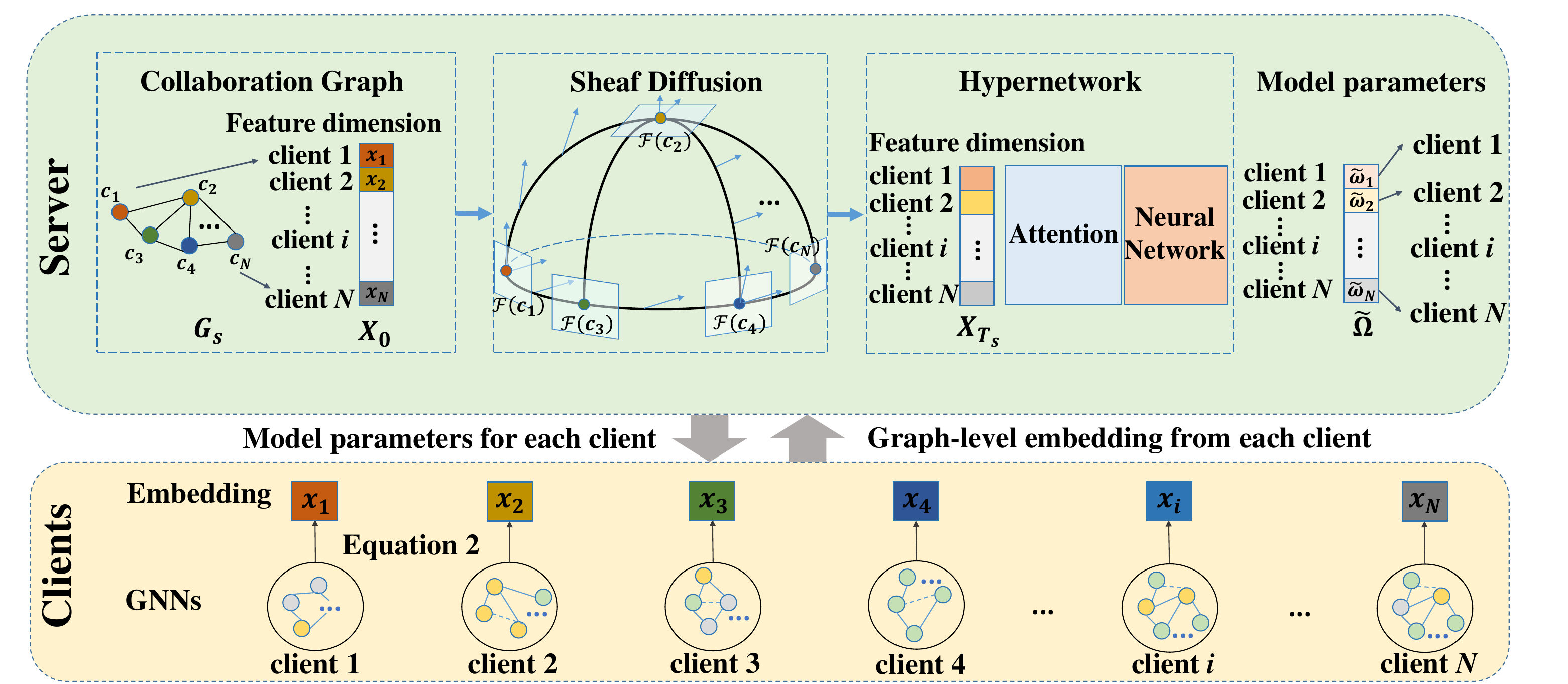} %width=0.85
\caption{The framework of FedSheafHN.}
\label{wholeframework}
\end{figure*}

In this work, to address the aforementioned challenges, we propose a novel personalized subgraph FL algorithm called Federated learning with Sheaf diffusion and HyperNetworks (FedSheafHN), as illustrated in \cref{wholeframework}. In the framework, the server constructs a collaboration graph where each client, possessing a local subgraph, is embedded at the graph level. This process is enhanced by sheaf diffusion, which improves the representation of clients' data by leveraging the underlying structure of the collaboration graph and aggregating information from other clients. This approach helps in understanding the complex inter-client relationships and ensures that the hypernetwork generates highly personalized models based on enriched client descriptions. Moreover, the hypernetwork is equipped with an attention layer to further aggregate cross-client information and is optimized for parallel operation across all clients. Through these techniques, FedSheafHN balances efficiency with effective personalization. This powerful framework adeptly aggregates information among clients’ graph-structured data and efficiently creates personalized models in FL settings that can be directly used by clients without additional training.
This not only enhances flexibility but also facilitates easy generalization to new clients. The experimental results demonstrate the superior performance of FedSheafHN over various baselines across multiple graph-structured datasets. Our main contributions are summarized as follows:
\begin{itemize}
\item We construct a collaboration graph for clients using their graph-level embeddings. This graph serves as a foundational basis for enhanced understanding and effective information aggregation across clients. Our data-driven client representations significantly boost the performance of HNs.
\item We apply sheaf diffusion to the server-constructed collaboration graph, which enhances the representation of clients and facilitates more effective aggregation of information across clients. 
\item We implement an attention-based hypernetwork and optimize it for parallel operation across all clients, efficiently generating highly personalized model parameters by leveraging enriched client descriptions from previous steps.
\item We empirically evaluate our model, demonstrating its superior performance over various baselines across multiple graph-structured datasets in heterogeneous subgraph FL scenarios. We also highlight its fast convergence and effective generalization to new clients.
\end{itemize}
%#####################################################################
\vspace{-2mm}
\section{Related work}
\vspace{-1mm}
\subsection{Federated Learning}
\vspace{-1mm}
\paragraph{FL.}
In the context of the distributed learning challenge, FL plays a pivotal role. 
FedAvg \cite{mcmahan2017communication} stands out as a notable strategy, involving local model training for each client and transmitting it to a central server for aggregation. 
To enhance the learning performance of local and global models or address heterogeneous local data, numerous FL methods have been designed, such as FedProx \cite{li2020federated}, MOON \cite{li2021model} and FedFM \cite{ye2023fedfm}.
Investigations also propose alternative methodologies, like distilling outputs \cite{chen2023spectral} or directly minimizing disparities in model outputs \cite{mohri2019agnostic}. 
%%%%%%%%%%%%%%%%%%%%%%%%%%%%%%%%%%%%%%%%%%%%%%%%%%%%%%%%%%%
\vspace{-2mm}
\paragraph{PFL.}
To address challenges related to data and device heterogeneity in FL, various PFL methods have been introduced, encompassing various approaches: local fine-tuning \cite{arivazhagan2019federated,wang2019federated,schneider2021personalization}, regularization for objective functions \cite{hanzely2020lower,hanzely2020federated,tan2022towards}, model mixing \cite{deng2020adaptive,mansour2020three,ma2022layer}, meta-learning \cite{jiang2019improving,fallah2020personalized,lee2024fedl2p} personalized parameter decomposition \cite{arivazhagan2019federated,bui2019federated,collins2021exploiting}, and differentially privacy \cite{agarwal2021skellam,noble2022differentially,li2024clients}, e.g. FedPer \cite{arivazhagan2019federated}, FedRep \cite{collins2021exploiting}, pFedMe \cite{tan2022towards}, pFedLA \cite{ma2022layer}, and FedL2P \cite{lee2024fedl2p}. 
Additionally, the concept of training multiple global models at the server has been explored for efficient PFL \cite{ghosh2020efficient,huang2021personalized}. This method involves training different models for distinct client groups, and clustering clients based on similarity.
Another strategy is to train individual client models collaboratively \cite{smith2017federated,zhang2020personalized,huang2021personalized,zhang2021parameterized}.
%%%%%%%%%%%%%%%%%%%%%%%%%%%%%%%%%%%%%%%%%%%%%%%%%%%%%%%%%%%%%%%
\vspace{-2mm}
\paragraph{Graph FL.} 
Recent research underscores the potential of integrating the FL framework into collaborative training of GNNs to uphold privacy \cite{he2021fedgraphnn,wang2022federatedscope}. This exploration falls into two primary categories: subgraph- and graph-level methods. Graph-level FL assumes clients possess entirely disjoint graphs, suitable for scenarios like molecular graphs. Studies \cite{xie2021federated,he2021spreadgnn,tan2023federated} delve into managing heterogeneity among non-IID graphs, where client graphs differ in labels.
In contrast to graph-level FL challenges, our focus centers on subgraph-level FL, introducing a unique challenge tied to the structure of graphs. Subgraphs, representing parts of a global graph, may have missing links between them. Existing methods \cite{wu2021fedgnn, zhang2021subgraph,yao2022fedgcn} address this by augmenting nodes and connecting them. However, this may compromise data privacy and increase communication overhead.
Another personalized subgraph FL \cite{baek2023personalized} addresses the missing link problem by exploring subgraph communities, representing densely connected clusters of subgraphs.
Unlike the above graph FL methods, our approach involves constructing the collaboration graph on the server to achieve heterogeneous client representation. This is considered as node-client two-level graph learning, which is beneficial for acquiring a diverse range of information.
%%%%%%%%%%%%%%%%%%%%%%%%%%%%%%%%%%%%%%%%%%%%%%%%%%%%%%%%%%%%%%%%
\vspace{-1mm}
\subsection{Hypernetworks}
\vspace{-1mm}
HNs \cite{klein2015dynamic} are deep neural networks designed to generate the weights for a target network responsible for the learning task. 
The underlying concept is that the output weights adapt based on the input provided to the hypernetwork \cite{klocek2019hypernetwork,navon2020learning}. 
SMASH \cite{brock2017smash} extended the HNs approach to facilitate Neural Architecture Search (NAS), encoding architectures as 3D tensors through a memory channel scheme. In contrast, our methodology involves encoding a network as a computation graph and deploying GNNs. While SMASH is tailored to predict a subset of weights, our model is designed to predict all free weights.
GHN \cite{zhang2018graph} is introduced as a search program for anytime prediction tasks, which not only optimizes for final speed but addresses the entire speed-accuracy trade-off curve, providing a comprehensive solution to the challenges.
pFedHN \cite{shamsian2021personalized} is presented by the use of HNs for FL with an input-task embedding. 
Furthermore, an improved method called hFedGHN \cite{xu2023heterogeneous} is presented to train heterogeneous graph-learning-based local models with GHN. 
HNs inherently lend themselves to the task of learning a diverse set of personalized models, given their ability to dynamically generate target networks conditioned on the input.
%####################################################################################
\vspace{-2mm}
\section{Methodology}
\vspace{-1mm}
\subsection{Problem Formulation}
\vspace{-1mm}
In the FL system, we envision a scenario with $\textit{N}$ clients, each denoted as client $i$, and a central server. Every client $i$ possesses a unique graph $G_{i}$, comprising the node set $M_{i}$ and edge set $E_{i}$, along with a corresponding feature matrix $V_{i}$ consisting of the node features $\set{v_{i,m} \given m\in M_{i}}$, where $v_{i,m}$ denotes the $m$-th node feature in client $i$. 
These clients independently design their models, denoted as $f(G_{i};\omega_{i})$, tailored to their specific tasks, where $\omega_{i}$ denotes the model parameters for client $i$.
In this context, each local graph $G_{i}$ can potentially be a subgraph of a larger, undisclosed global graph $G$. 

Our approach aims to enhance the performance of the personalized model $f(G_{i};\omega_{i})$ on each client through graph FL. To facilitate this, FedSheafHN constructs a collaboration graph, denoted as $G_{s}$, at the server. The framework also includes a sheaf diffusion model $S(\cdot; \theta)$ and a hypernetwork denoted as $H(\cdot; \varphi)$, where $\theta$ and $\varphi$ are the corresponding model parameters, respectively. 

The complete parameter set $\Omega$, which includes $\omega_{i}$ for each client, consists of the generated parameter $\tilde{\Omega}$ and the learnable parameter $\Omega'$. These correspond to the parameter sets $\{\tilde{\omega}_i\}$ and $\{\omega'_i\}$, respectively. Each $\omega_i$ includes both $\tilde{\omega}_i$ and $\omega'_i$.
As a result, we formulate the training objective as follows:
\begin{align}
\label{loss_func}
\argmin_{\varphi, \theta, \Omega} \mathcal{L}(\Omega),
\end{align}
where $\mathcal{L}(\cdot)$ denotes the set of loss functions for clients, i.e., $\mathcal{L} = \set{\mathcal{L}_i \given i = 1, 2, \dots, N}$. $\mathcal{L}_{i}$ is the loss function for client $i$. 
%######################################################################################
\vspace{-1mm}
\subsection{Client Representation}
\vspace{-1mm}
\paragraph{Collaboration Graph Construction.}  
In subgraph FL systems, a key challenge for the server is the inability to fully perceive the overarching graph structure, complicating the determination of the most effective collaboration methods among clients. Consequently, our objective is to devise an approach that adeptly unravels the intrinsic geometric relationships within clients data. To this end, we formulate a novel collaboration graph with client data-driven features to map the interactions among clients. This graph not only facilitates understanding of client collaboration but also serves as the foundation for enhancing the representation of client-specific data.

Let $v_{i,m}^{(T_c)}$ be the updated embedding vector of node $m$ in client $i$ after $T_c$ steps of the client model $f(G_i;\omega_{i})$ propagation. The graph-level embedding of client $i$ is given by
\begin{align}
\label{graph_embedding}
x_{i}=\frac{1}{\left | M_{i} \right |} \sum_{m\in M_{i}}  v_{i,m}^{(T_c)}. 
\end{align}
The graph-level embedding $x_{i}$ for each client $i$ is transmitted to the server and serves as the feature representation of that client. We conceptualize a collaboration graph $G_{s}$ on the server side, which encompasses all $\textit{N}$ clients. This graph is defined with a node set $\{c_{1}, c_{2},\dots ,c_{N} \}$, representing each client. For each node $c_{i}$ in $G_{s}$, the corresponding graph-level embedding $x_{i}$ is used as its feature. As an initialization, $G_{s}$ is set to be a complete graph to allow for every pairwise client relationship. The initial feature matrix $X_{0}$ consists of $x_{i}$, i.e., $X_{0} = (x_i)_{i=1,\dots,N}$.
%%%%%%%%%%%%%%%%%%%%%%%%%%%%%%%%%%%%%%%%%%%%%%%%%%%%%%%%
\vspace{-2mm}
\paragraph{Graphical Representation Learning.}  
To identify latent inter-client relationships that are not immediately apparent and enrich client descriptions by aggregating information from other clients, our method incorporates elements of (cellular) sheaf theory. A cellular sheaf assigns a vector space to every node and edge in the graph, establishing a linear map for every incident node-edge pair \cite{curry2014sheaves, hansen2020sheaf, bodnar2022neural}. 
These vector spaces are associated with points on a manifold, with the sheaf Laplacian describing the transport of elements via rotations in neighboring vector spaces in discrete settings, where the graph plays the role of the manifold \cite{bodnar2022neural}. Inspired from this, our framework employs the server-side collaboration graph instead as the manifold, forming linear maps between client pairs to capture the collaboration graph's underlying geometry and client interconnections. 

Mathematically, a cellular sheaf $(G,\mathcal{F})$ on an undirected graph $G = (M,E)$ includes three components:
\textit{(a)} A vector space $\mathcal{F}(m)$ ($m \in M $), \textit{(b)} A vector space $\mathcal{F}(e)$ ($e\in E $), \textit{(c)} A linear map $\mathcal{F}_{m\unlhd e}:\mathcal{F}{(m)}\to \mathcal{F}{(e)}$ for each incident $m\unlhd e$ node-edge pair.
The sheaf Laplacian of a sheaf $(G,\mathcal{F})$ is a linear map given by
\begin{align}
\!L_{\mathcal{F}}(X)_{m} :=  {\textstyle \sum_{m,n\unlhd e}^{}}\mathcal{F}_{m\unlhd e}^{\top }(\mathcal{F}_{m\unlhd e}X_{m}- \mathcal{F}_{n\unlhd e}X_{n}), 
\end{align}
Let $D$ be the block-diagonal of $\mathit{L}_{\mathcal{F}}$, the normalised sheaf Laplacian is given by
\begin{align}
\Delta_{\mathcal{F}}: = \mathit{D}^{-1/2} \mathit{L}_{\mathcal{F}} \mathit{D}^{-1/2}. 
\end{align}
This operator can be used to describe the process of vector space elements being transported and rotated into neighboring vector spaces, capturing the essence of element transitions within the structure of the sheaf. A diffusion-type model \cite{bodnar2022neural} is introduced from Equation \cref{sheaf_diffusion}, which enables end-to-end learning of the underlying sheaf directly from data as follows
\begin{align}
\label{sheaf_diffusion}
\!X_{t}=X_{t-1}-\!\sigma \left ( \Delta_{\mathcal{F}^{(t-1)}} (\mathrm{I} \otimes \mathrm{W}_{1}^{(t-1)} ) X_{t-1}\mathrm{W}_{2}^{(t-1)}\right ),
\end{align}
where $\sigma$ denotes a non-linearity function, $\mathrm{I}$ is the identity matrix, $\otimes$ is the Kronecker product, $\mathrm{W}_{1}^{(t-1)}$ and $\mathrm{W}_{2}^{(t-1)}$ are learnable weight matrices. The feature of the updated collaboration graph at the $t$-th step is denoted as $X_{t}\in \mathbb{R}^{N\times d}$.

In our framework, the neural sheaf diffusion model is utilized to adjust the geometry of the collaboration graph dynamically, ensuring precise representation and effective knowledge integration across clients. Let $X_{0}$ be the input for the neural sheaf diffusion model with $T_s$ integration time steps. Then, we have 
\begin{align}
X_{T_s} = S(X_{0}; \theta).  
\end{align}

The collaboration graph $G_{s}$ also evolves in tandem with the client model training, with graph-level embedding vectors progressively refined for better representation of client graph features. Integrated seamlessly into existing client model training, this process eliminates the need for an additional model and aligns the evolution of $G_{s}$ with ongoing learning and adaptation in client models.
%#############################################################################################
\vspace{-1mm}
\subsection{Hypernetworks Optimization}
\vspace{-1mm}
To generate personalized model parameters for each client, we utilize HNs that take input from client embedding vectors, which have been updated with information from other clients in previous operations. 
Unlike conventional HNs that use a basic multilayer perceptron (MLP) \cite{zhang2018graph, shamsian2021personalized, xu2023heterogeneous} and might ignore client relationships, our framework is tailored to integrate potential latent relationships between the elements of each client's embedding vector, ensuring in-depth analysis within the network. Furthermore, we optimize the HNs for parallel operation across all clients by leveraging the feature matrix of the collaboration graph.

To implement this, we incorporate an attention layer on top of the MLP into the HNs, where the attention-based embedding is given by
\begin{align}
\label{hn_att}
X_{T_s}^{\mathrm{att}}=f_\mathrm{att}(X_{T_s}),
\end{align}
where $f_\mathrm{att}(\cdot)$ denotes the attention operation, which provides an attention matrix ${A_{\mathrm{att}}}\in \mathbb{R}^{N\times N}$ for $X_{T_s}$.
This allows the model to focus more effectively on important elements in the clients' embedding vectors and make good use of the enriched graph representations.

The partial parameter $\tilde \Omega$ for the model is generated from the hypernetwork by
\begin{align}
\label{eq.HN_Omega}
\tilde \Omega = H(X_{T_s}; \varphi).
\end{align}
The hypernetwork $H$ is tasked with learning a set of personalized models, which, together with the learned parameter $\Omega'$, form the complete parameter $\Omega$ of the model.
Here, $\Omega$ denotes the model weights of clients and is structured so that each row $i$ corresponds to the model weights $\omega_{i}$ of client $i$.
%################################################################################################
\vspace{-1mm}
\subsection{Learning Procedure}
\vspace{-1mm}
The parameter $\Omega$ is based on the produced $\tilde \Omega$ and the learned $\Omega'$.
Specifically, the parameter $\tilde \Omega$ based on $\theta$ and $\varphi$ is updated on the server, and the parameter $\Omega'$ is updated on the clients.
From \cref{loss_func}, we calculate the gradients of FedSheafHN for the parameters $\theta, \varphi$ on the server using the chain rule as follows
\begin{align}
\label{chain_rule}
\nabla _{\theta, \varphi }\mathcal{L}(\tilde \Omega) =\nabla _{\tilde \Omega}\mathcal{L}(\tilde \Omega)\cdot\nabla _{\theta, \varphi }\tilde \Omega,
\end{align}
from which the corresponding gradients with respect to $\theta, \varphi$ are given by 
\begin{align}
\label{gd_s}
\nabla_{\theta}\tilde{\Omega} =\nabla_{X_{T_s}}H(X_{T_s};\varphi)\cdot \nabla_{\theta} S(X_0; \theta),
\end{align}
as well as 
\begin{align}
\label{gd_h}
\nabla _{\varphi }\tilde{\Omega} = \nabla_{\varphi}H(X_{T_s};\varphi ).
\end{align}

Once the generated model parameter $\Omega$ for the current round is obtained, each client performs $T_c$ rounds of local training to obtain the updated model $\Omega_{T_c}$ including $\tilde\Omega_{T_c}$. We use a more general update rule to get $\nabla _{\tilde \Omega}\mathcal{L}(\tilde \Omega):= \tilde\Omega_{T_c}-\tilde\Omega$ from the clients. Throughout the training process, the models $S$ and $H$ are learned jointly based on \cref{gd_s} and \cref{gd_h}.

The FedSheafHN algorithm, outlined in Algorithm \ref{fedsheafhn}, commences with the server constructing a collaboration graph using the clients' graph-level embeddings. This collaboration graph evolves dynamically throughout the training process. Subsequently, the server employs a neural sheaf diffusion model to aggregate and update client information in the embedding vectors. 
An advanced hypernetwork generates client model parameters from these updated vectors. Clients then utilize these models for local training on their respective datasets for several epochs and transmit back model updates to the server. The server optimizes models $S$ and $H$ based on \cref{gd_s} and \cref{gd_h}, iteratively refining the learning process. 

%##############################################
\begin{algorithm}[!htb]
\caption{FedSheafHN}
\label{fedsheafhn}

\begin{algorithmic}[1]
    \State {\bfseries Input:} Total round $R$, client number $N$, client local training round $T_c$
\end{algorithmic}
\begin{minipage}[t]{0.58\textwidth}
\begin{algorithmic}[1]
    \State {\bfseries Server:}
    \For{communication round in $\{1,2,\dots,R\}$}
        \If{update embedding vector}
            \State construct $G_{s}$ with feature matrix $X_{0}$ %collaboration graph
        \EndIf
        \State update client embedding $X_{T_s}=S(G_{s};\theta)$
        \State generate model parameter $\tilde \Omega=H(X_{T_s};\varphi)$ and distribute to $\tilde\omega_{1}, \tilde\omega_{2}, \dots, \tilde\omega_{N}$
        \For{client $i$ in $\{1,2,\dots, N\}$}
            \State Client Update ($i, \tilde\omega_{i}$)
        \EndFor
        \State update model $S$,$H$
    \EndFor
\end{algorithmic}
\end{minipage}
\hfill
\begin{minipage}[t]{0.4\textwidth}
\begin{algorithmic}[1]
    \State {\bfseries Client Update}($i, \tilde\omega_{i}$):
    \For{local epoch in $\{1,2,\dots, T_c\}$}
        \State train on generated model $f(G_{i};\tilde\omega_{i};\omega'_{i})$
    \EndFor
    \If{update embedding vector}
        \State generate graph level embedding as \cref{graph_embedding}
    \EndIf
    \State return $\Delta \tilde\omega_{i}:= \tilde\omega_{i}^{(T_{c})}-\tilde\omega_{i}$
\end{algorithmic}
\end{minipage}
\end{algorithm}
%#######################################################################################
\vspace{-4mm}
\section{Experiments}
\vspace{-3.5mm}
We assess the performance of FedSheafHN on six datasets, focusing on node classification tasks within two distinct subgraph FL scenarios. To measure the performance of methods, we utilize Federated Accuracy, defined as $\frac{1}{N} \sum_{i\in N} \mathrm{Acc}(f(G_{i};\omega_{i}))$, where $\mathrm{Acc}(\cdot)$ denotes the accuracy of its argument model.  
\vspace{-3mm}
\subsection{Experimental settings}
\vspace{-1mm}
\paragraph{Datasets.}  
In accordance with the experimental framework outlined in \cite{zhang2021subgraph, baek2023personalized}, we meticulously partitioned datasets into distinct segments, assigning each client in the FL process a dedicated subgraph. This deliberate allocation ensures that each participant manages a segment of a larger, original graph. The experimentation covered six datasets: Cora, CiteSeer \cite{sen2008collective}, Pubmed, ogbn-arxiv \cite{hu2020open}, Computer, and Photo \cite{mcauley2015image, shchur2018pitfalls}. The partitioning of these datasets was executed through the METIS graph partitioning algorithm \cite{karypis1997metis}, with the number of subsets predetermined. Our experiments embraced both non-overlapping and overlapping node scenarios. In the non-overlapping node scenario, the METIS output was directly utilized, creating distinct subgraphs without shared nodes, thus cultivating a more heterogeneous setting. The statistical characteristics of these datasets in the non-overlapping scenario are meticulously presented in Table~\ref{nonoverlapping_dataset_statistics}. In the overlapping node scenario, Table~\ref{overlapping_dataset_statistics}, subgraphs featured shared nodes, achieved by sampling smaller subgraphs from the initial METIS partitioned results (see additional details in Appendix~\ref{app_datasets}). 
%%%%%%%%%%%%%%%%%%%%%%%%%%%%%%%%%%%%%%%%%%%%%%%%%
\vspace{-2mm}
\paragraph{Baselines.}  
\textbf{1) Local} : clients train models locally. 
\textbf{2) FedAvg} \cite{mcmahan2017communication}: A FL baseline which involves the server aggregating client models that have been updated locally based on the volume of training data per client. 
\textbf{3) FedPer} \cite{arivazhagan2019federated}: A PFL model where clients share only the base layers of their models, maintaining personalized layers locally for tailored learning.
\textbf{4) FedSage+} \cite{zhang2021subgraph}: A subgraph FL baseline which addresses missing links in local subgraphs by training a missing neighbor generator. 
\textbf{5) pFedHN} \cite{shamsian2021personalized}: A PFL baseline which uses a central hypernetwork to generate personalized models for each client. 
\textbf{6) pFedGraph} \cite{ye2023personalized}: A PFL baseline which apply a collaboration graph to tailor interactions among clients based on model similarity and dataset size. 
\textbf{7) FED-PUB} \cite{baek2023personalized}: A personalized subgraph FL baseline which utilize similarity matching and weight masking techniques to enhance model personalization.
\textbf{8) FedSheafHN}: Our proposed personalized subgraph FL framework that uses a collaboration graph enhanced by sheaf diffusion to generate highly personalized models with optimized hypernetworks.
See the detailed introduction of baselines in Appendix~\ref{app_baselines}.
%%%%%%%%%%%%%%%%%%%%%%%%%%%%%%%%%%%%%%%%%%%%%%%%%%%%%%%%
\vspace{-2mm}
\paragraph{Implementation Details.} 
\label{implementation}
The client models are two-layer GCNs for all baselines. For our framework, the sheaf diffusion model utilized is Diag-NSD \cite{bodnar2022neural}, which allows learning fewer parameters per edge. The HNs in our framework comprises an attention layer, two hidden layers, and an output layer. The hidden dimension of GCNs and HNs is set to 128. The Adam optimizer \cite{kingma2014adam} is applied for optimization. We randomly allocate 40\% of nodes for training, 30\% for validation, and another 30\% for testing across all datasets. We perform FL over 100 communication rounds for the Cora, CiteSeer, Pubmed, Computer and Photo datasets. For the ogbn-arxiv dataset, we extended this to 200 rounds. The number of local training epochs varied between 1 and 15, tailored to each dataset.
%%%%%%%%%%%%%%%%%%%%%%%%%%%%%%%%%%%%%%%%%%%%%%%%%%%%%%%%%
\begin{table}[!t] %[!htb]
\caption{Results on the non-overlapping node scenario. The reported results are mean and standard deviation over five different runs. The best and the second-best results are highlighted in bold and underlined, respectively.}
\label{non-overlapping}
\vspace{-5mm}
\begin{center}
\tiny
\resizebox{\textwidth}{!}{
\begin{tabular}{lccccccc}
\toprule
\multirow{2}{*}{Methods} & \multicolumn{2}{c}{Cora}            & \multicolumn{2}{c}{Citeseer}        & \multicolumn{2}{c}{Pubmed}   \\ 
\cline{2-8} 
                & 10 Clients & 20 Clients & 10 Clients & 20 Clients & 10 Clients & 20 Clients\\ 
\hline
Local           & 79.94$\pm$0.24 & 80.30$\pm$0.25 & 67.82$\pm$0.13 & 65.98$\pm$0.17 & 82.81$\pm$0.39 & 82.65$\pm$0.03  &\\ \hline
FedAvg          & 72.38$\pm$2.45  & 69.81$\pm$13.28  & 65.71$\pm$0.37  & 63.08$\pm$7.38 & 79.88$\pm$0.07  & 78.48$\pm$7.65 &\\
FedPer        & 79.35$\pm$0.04  & 78.01$\pm$0.32  & 70.53$\pm$0.28  & 66.64$\pm$0.27   & 84.20$\pm$0.28  & 84.72$\pm$0.31   & \\
FedSage+       & 69.05$\pm$1.59 & 57.97$\pm$ 12.6 & 65.63$\pm$3.10  & 65.46$\pm$0.74  & 82.62$\pm$0.31  & 80.82$\pm$0.25     & \\
pFedHN          & 65.24$\pm$0.21  & 70.65$\pm$2.21  & 63.45$\pm$0.44  & 58.98$\pm$1.60  & 70.46$\pm$0.17  & 71.24$\pm$2.50     & \\
pFedGraph       & 75.52$\pm$0.94  & 74.61$\pm$0.51  & 71.61$\pm$0.64 & 67.43$\pm$0.49  & 80.38$\pm$0.30  & 80.48$\pm$0.58     & \\
FED-PUB         & \udcloser{81.54$\pm$0.12} & \udcloser{81.75$\pm$0.56}  & \udcloser{72.35$\pm$0.53}  & \udcloser{67.62$\pm$0.12}   & \udcloser{86.28$\pm$0.18}  & \udcloser{85.53$\pm$0.30}   &\\
FedSheafHN(Ours)& \textbf{83.49$\pm$0.22}  & \textbf{82.35$\pm$0.18}  & \textbf{75.27$\pm$0.39}  & \textbf{72.20$\pm$0.24}   & \textbf{86.50$\pm$0.09}  & \textbf{85.57$\pm$0.05}     \\ \bottomrule
\multicolumn{8}{l}{}
\vspace{-2mm}
\\ \toprule
\multirow{2}{*}{Methods} & \multicolumn{2}{c}{Amazon-Computer} & \multicolumn{2}{c}{Amazon-Photo}    & \multicolumn{2}{c}{ogbn-arxiv}        \\ \cline{2-8} 
                & 10 Clients & 20 Clients & 10 Clients & 20 Clients & 10 Clients & 20 Clients   &\\ \hline
Local           & 88.91$\pm$0.17 & 89.52$\pm$0.20  &  91.80$\pm$0.02  &  90.47$\pm$0.15& 64.92$\pm$0.09 &  65.06$\pm$0.05&  \\ 
\hline
FedAvg          & 66.78$\pm$0.00  &  71.44$\pm$0.08  &  79.61$\pm$3.12  &  82.12$\pm$0.02    &  48.77$\pm$2.88   &     42.02$\pm$17.09  \\
FedPer          & 89.73$\pm$0.04  & 87.86$\pm$0.43  & 91.76$\pm$0.23  & 90.59$\pm$0.06   & 64.99$\pm$0.18  & 64.66$\pm$0.11   & \\
FedSage+        & 80.50$\pm$1.30   & 70.42$\pm$0.85  & 76.81$\pm$8.24  & 80.58$\pm$1.15  & 64.52$\pm$0.14  & 63.31$\pm$0.20     & \\
pFedHN          & 66.85$\pm$0.09  & 69.94$\pm$1.27  & 74.12$\pm$0.90  & 79.90$\pm$2.25  & 48.55$\pm$0.59  & 47.64$\pm$0.40     & \\
pFedGraph       & 66.45$\pm$0.83  & 71.57$\pm$0.36  & 74.57$\pm$1.05 & 84.04$\pm$0.50  & 56.37$\pm$0.31  & 56.19$\pm$0.83     & \\
FED-PUB         &\udcloser{90.55$\pm$0.13} & \udcloser{90.12$\pm$0.09}  & \udcloser{92.73$\pm$0.18}  & \udcloser{91.92$\pm$0.12}     &\udcloser{66.58$\pm$0.08}           &\udcloser{66.64$\pm$0.12}  &\\
FedSheafHN(Ours) & \textbf{90.56$\pm$0.03} & \textbf{91.00$\pm$0.09} & \textbf{94.22$\pm$0.10}  & \textbf{92.99$\pm$0.05}   & \textbf{71.28$\pm$0.03}  & \textbf{71.75$\pm$0.09} &\\ 
\bottomrule
\end{tabular}}
\vspace{-2mm}
\end{center}
\end{table}
%%%%%%%%%%%%%%%%%%%%%%%%%%%%
\begin{figure}[!t]
\centering
\vspace{-3mm}
\includegraphics[width=\textwidth]{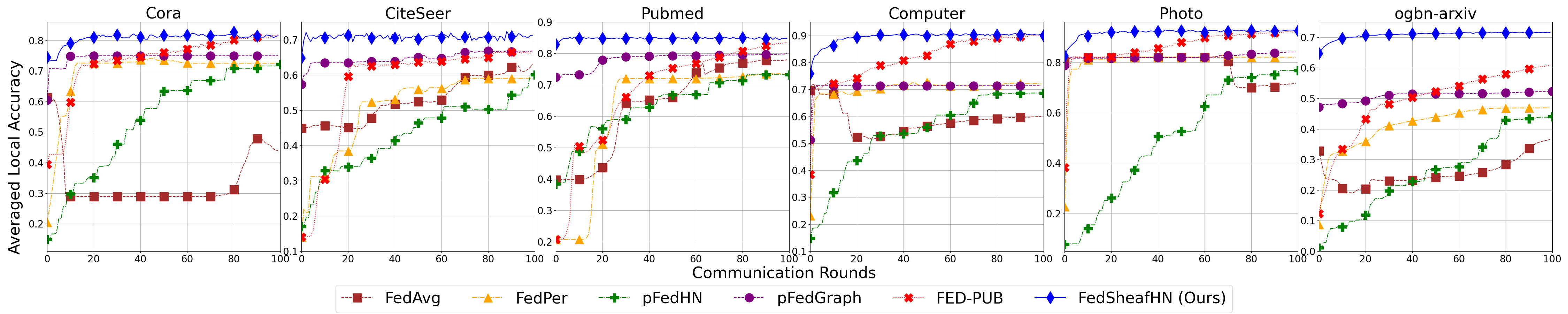} 
\vspace{-6mm}
\caption{Convergence plots for the non-overlapping scenario with 20 clients.}
\vspace{-5mm}
\label{conv_ds}
\end{figure}
%%%%%%%%%%%%%%%%%%%%%%%%%%%
%%%%%%%%%%%%%%%%%%%%%%%%%%%%%%%%%%%%%%%%%%%%%%%%%%%%%%%%%
\vspace{-3mm}
\subsection{Experiment Results}
\vspace{-2mm}
\paragraph{Results of Non-overlapping Scenario.}
In \cref{non-overlapping}, we present the node classification results for the non-overlapping scenario, characterized by a notably heterogeneous subgraph FL challenge. Our proposed method, FedSheafHN, stands out as it consistently outperforms other baseline approaches. 
FED-PUB, while achieving commendable results through its identification of community structures via similarity estimation and selective filtering of irrelevant weights from diverse communities, encounters limitations in scenarios with many distinct clients. In these complex cases marked by extreme heterogeneity among clients, solely relying on similarity measures proves insufficient for adequately inferring client relationships.
FedSheafHN addresses this challenge to some extent by enhancing the representation of relationships among clients in heterogeneous scenarios. 
Its innovative approach improves relationship inference in complex cases, enhancing feature representation in highly heterogeneous scenarios and thus boosting the performance of HNs in creating personalized client models.
The method's adaptability to diverse and challenging situations positions FedSheafHN as a promising choice for addressing the complexities inherent in non-overlapping scenarios.
%%%%%%%%%%%%%%%%%%%%%%%%%%%%%%%%%%%%%%%%%%%%%%%%%%%%%%%%
\vspace{-3mm}
\paragraph{Results of Overlapping Scenario.} 
We investigated the overlapping scenario, characterized by lower heterogeneity as outlined in \cref{overlapping}. Despite the reduced heterogeneity, our method exhibits advantages across various datasets, affirming its efficacy. In contrast to the non-overlapping scenario, our method's efficiency is not as pronounced in the overlapping counterpart. This discrepancy may stem from the less pronounced heterogeneity within the overlapping scenario compared to the non-overlapping counterpart. The diverse feature representation employed by our method is tailored to address the intricacies of this specific scenario.

\begin{figure}[!t] %H
\centering
\includegraphics[width=\textwidth]{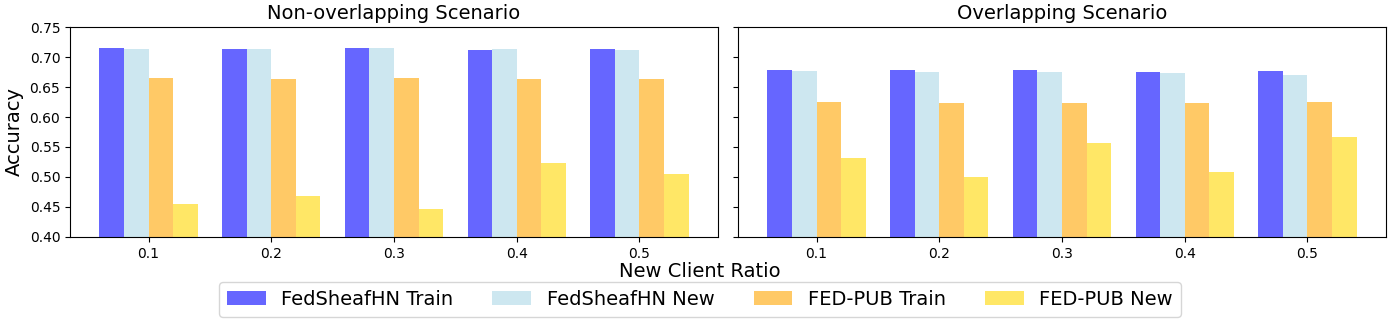}
\vspace{-6mm}
\caption{Results of generalization to new clients on the ogbn-arxiv dataset. ``Train'' represents the average test accuracy for trained clients, while ``New'' indicates the average test accuracy for newly joined clients.}
\label{newclients}
\vspace{-8mm}
\end{figure}

Nonetheless, it is worth emphasizing that our methodology demonstrates superior performance in comparison to other contemporary state-of-the-art benchmarks. Despite potential challenges posed by overlapping scenarios, the capacity of our approach to surpass current methodologies underscores its robustness and versatility. This illustrates the efficacy of our approach in negotiating diverse degrees of heterogeneity, thereby positioning it as a more dependable option.
%%%%%%%%%%%%%%%%%%%%%%%%%%%%%%%%%%%%%%%%%%%%%%%%%%%%%%%%%
\begin{table*}[!t] %[!htb]
\caption{Results on the overlapping node scenario. The reported results are mean and standard deviation over five different runs. The best and the second-best results are highlighted in bold and underlined, respectively.}
\label{overlapping}
\vspace{-2mm}
\begin{center}
\tiny
\resizebox{\textwidth}{!}{
\begin{tabular}{lccccccc}
\toprule
\multirow{2}{*}{Methods} & \multicolumn{2}{c}{Cora}   & \multicolumn{2}{c}{Citeseer}   & \multicolumn{2}{c}{Pubmed}  \\ \cline{2-8} 
                & 30 Clients & 50 Clients & 30 Clients & 50 Clients & 30 Clients & 50 Clients   \\ \hline
Local           & 71.65$\pm$0.12  & 76.63$\pm$0.10 & 64.54$\pm$0.42  & 66.68$\pm$0.44 & 80.72$\pm$0.16  & 80.54$\pm$0.11    \\ \hline
FedAvg          & 63.84$\pm$2.57  & 57.98$\pm$0.06  & 66.11$\pm$1.50  & 58.00$\pm$ 0.29 & 83.11$\pm$0.03  & 82.24$\pm$0.73      \\
FedPer        & 74.18$\pm$0.24  & 74.42$\pm$0.37  & 65.19$\pm$0.81  & 67.64$\pm$0.44  & 70.08$\pm$0.38  & 71.13$\pm$0.04      \\
FedSage+        & 51.99$\pm$0.42  & 55.48$\pm$11.5  & 65.97$\pm$0.02  & 65.93$\pm$0.30  & 82.14$\pm$0.11  & 80.31$\pm$0.68     & \\
pFedHN          & 48.71$\pm$2.19  & 49.19$\pm$2.54  & 54.67$\pm$1.28  & 46.34$\pm$2.24  & 66.00$\pm$2.22  & 63.55$\pm$1.35     & \\
pFedGraph       & \udcloser{77.72$\pm$0.41}  & 77.69$\pm$0.20  & \udcloser{69.60$\pm$0.11}  & 67.84$\pm$0.75  & 83.12$\pm$0.37  & 82.60$\pm$0.27     & \\
FED-PUB         & 75.40$\pm$0.54  & \udcloser{77.84$\pm$0.23}  & 68.33$\pm$0.45  & \textbf{69.21$\pm$0.30}  & \textbf{85.16$\pm$0.10}  & \textbf{84.84$\pm$0.12}      \\
FedSheafHN(Ours)& \textbf{80.30$\pm$0.11}  & \textbf{78.06$\pm$0.22}  & \textbf{71.90$\pm$0.15}  & \udcloser{68.59$\pm$0.12}   & \udcloser{84.45$\pm$0.03}  & \udcloser{83.65$\pm$0.02}    \\ \bottomrule
\multicolumn{8}{l}{} 
\vspace{-2mm}
\\ \toprule
\multirow{2}{*}{Methods} & \multicolumn{2}{c}{Amazon-Computer} & \multicolumn{2}{c}{Amazon-Photo}    & \multicolumn{2}{c}{ogbn-arxiv}        \\ 
                \cline{2-8} 
                & 30 Clients & 50 Clients & 30 Clients & 50 Clients & 30 Clients & 50 Clients  \\ \hline
Local           & 86.66$\pm$0.00 &  87.04$\pm$0.02    &   90.16$\pm$0.12  & 90.42$\pm$0.15   &  61.32$\pm$0.04 &60.04$\pm$0.04      \\ \hline
FedAvg          & 68.99$\pm$3.97 &  67.69$\pm$0.00    &  83.74$\pm$0.72 &  75.59$\pm$0.06   &   49.91$\pm$2.84  & 52.26$\pm$1.02\\
FedPer        & 87.99$\pm$0.23  & 88.22$\pm$0.27  & 91.23$\pm$0.16  & 90.92$\pm$0.38  & 62.29$\pm$0.04  & 61.24$\pm$0.11     \\
FedSage+        & 81.33$\pm$1.20  & 76.72$\pm$0.39  & 88.69$\pm$0.99  & 72.41$\pm$1.36  & 59.90$\pm$0.12  & 60.95$\pm$0.09     & \\
pFedHN          & 53.90$\pm$1.91  & 47.63$\pm$0.94  & 60.37$\pm$2.70  & 57.57$\pm$3.60  & 41.08$\pm$1.02  & 37.52$\pm$1.15     & \\
pFedGraph       & 76.77$\pm$1.53  & 74.46$\pm$0.42  & 77.07$\pm$0.69  & 87.81$\pm$1.96  & 59.29$\pm$0.39  & 58.96$\pm$0.56     & \\
FED-PUB         & \udcloser{89.15$\pm$0.06} &  \udcloser{88.76$\pm$0.14}    & \udcloser{92.01$\pm$0.07} &  \udcloser{91.71$\pm$0.11}  & \udcloser{63.34$\pm$0.12} &  \udcloser{62.55$\pm$0.12}   \\ 
FedSheafHN(Ours)& \textbf{89.25$\pm$0.02} & \textbf{89.29$\pm$0.05}  & \textbf{93.34$\pm$0.04}  & \textbf{92.36$\pm$0.06}   & \textbf{67.69$\pm$0.10}  & \textbf{67.51$\pm$0.05}   & \\ \bottomrule
\end{tabular}
}
\vspace{-2mm}
\end{center}
\end{table*}
%%%%%%%%%%%%%%%%%%%%%%%%%%%%%%%
\begin{figure}[!t]
\centering
\vspace{-3mm}%4
\includegraphics[width=\textwidth]{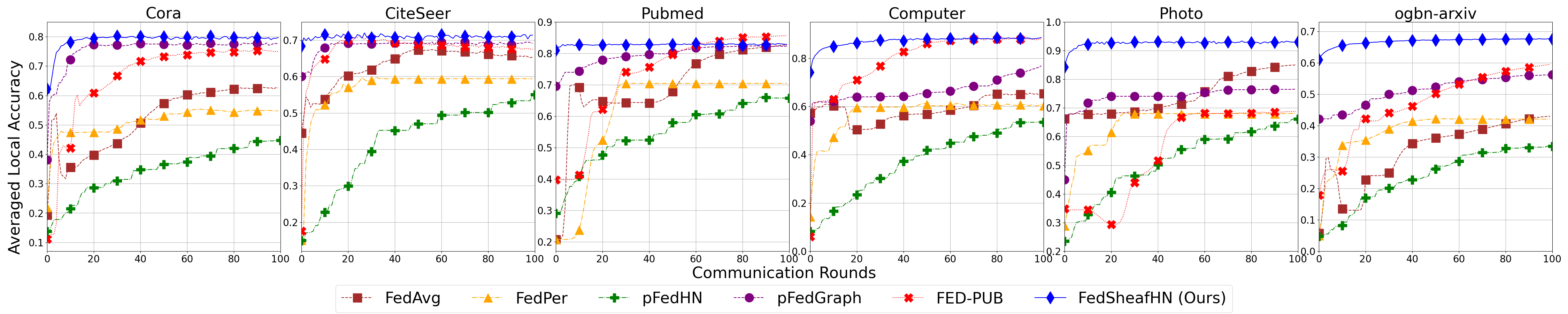}
\vspace{-6mm}
\caption{Convergence plots for the overlapping scenario with 30 clients.}
\vspace{-5mm}%8
\label{conv_ov}
\end{figure}
%%%%%%%%%%%%%%%%%%%%%%%%%%%%%%%%%%%%%%%%%%%%%%%%%%%%%%%%%%%%%%%%%%%%
\vspace{-3mm}
\paragraph{Fast Convergence.}   
\Cref{conv_ds} and \Cref{conv_ov} illustrate that FedSheafHN achieves rapid convergence. This efficiency is likely attributed to our framework's ability to discern underlying client cooperation and facilitate intelligent information sharing. Additionally, the hypernetwork, employing learned attention mechanisms and operating in parallel across all clients, efficiently generates model parameters and enhances the integration of global information.
%%%%%%%%%%%%%%%%%%%%%%%%%%%%%%%%%%%%%%%%%%%%%%%%%%%%%%%%%%%%%%%%%%%%%
\vspace{-3mm}
\paragraph{Generalization to New Clients.} 
We are also interested in evaluating the performance of FedSheafHN on clients not encountered during training. In typical scenarios where models are shared among clients, this would usually require retraining or fine-tuning the shared model. However, with FedSheafHN, once the shared model $S(\cdot; \theta)$ and the hypernetwork $H(\cdot; \varphi)$ are trained on one group of clients, expanding to new clients requires minimal effort. We simply freeze the weights of both the sheaf model and the hypernetwork ($\theta$ and $\varphi$), then initialize the new client's model. The only primary task is to set up and train the new client's model over several local epochs to derive a graph-level embedding ($x_{new}$). To evaluate FedSheafHN in this setting, we use the ogbn-arxiv dataset with 20 clients in the non-overlapping scenario and 30 clients in the overlapping scenario. Clients are divided into training or new groups based on new client ratios of 0.1, 0.2, 0.3, 0.4, and 0.5. The results, presented in \Cref{newclients}, show that FedSheafHN generates models for new clients with accuracy comparable to those of trained clients, only incurring a slight performance reduction. Furthermore, it requires just one communication round to obtain the graph-level embedding from new clients.
%%%%%%%%%%%%%%%%%%%%%%%%%%%%%%%%%%%%%%%%%%%%%%%%%%%%%%%%%%%%%%%%%%%%
\vspace{-3mm}
\paragraph{Ablation Study.}
As demonstrated in \cref{ab_ogbn_table}, our ablation study sequentially added each component to the base model (equivalent to FedAvg) to assess their impact. The results illuminate the positive influence of each element: constructing a collaboration graph using client graph-level embedding (collaboration graph), clients dynamically updating these embeddings to server along the training (dynamic embedding), applying sheaf diffusion to enhance the collaboration graph (sheaf diffusion), an improved hypernetwork working in parallel across all clients (hypernetwork), and incorporating an attention layer into the enhanced hypernetwork (attention).

The integration of these components results in a substantial overall improvement in performance. Nevertheless, the extent of contribution varies, with constructing the collaboration graph and using the improved hypernetwork being particularly impactful. In the ``+ Hypernetwork'' variant using a one-hot vector for each client, performance noticeably declines compared to the ``+ Collaboration graph''.
%%%%%%%%%%%%%%%%%%%%%%%%%%%%%%%%%%%%%%%%%
\vspace{-2mm}
\noindent 
\begin{minipage}[!t]{0.58\textwidth} 
    \centering
    \scriptsize
    \captionof{table}{Ablation studies of proposed FedSheafHN on the ogbn-arxiv dataset.}
    \label{ab_ogbn_table}
    \resizebox{\columnwidth}{!}{%
        \begin{tabular}{lcc}
        \toprule
        \multirow{2}{*}{Methods} & non-overlapping &overlapping \\ 
                         & 20 clients   & 30 clients   \\ 
        \midrule
        Base model (FedAvg)                    & 36.56  & 42.95    \\
        + Hypernetwork                         & 61.26  & 56.53      \\
        + Collaboration graph                  & 68.51  & 64.68  \\
        + Sheaf diffusion                      & 70.05  & 65.87   \\
        + Attention                            & 70.97  & 67.09   \\
        + Dynamic embedding (Ours)             & 71.87  & 67.89   \\ 
        \bottomrule
        \end{tabular}
        }
    \vspace{-2.5mm} %-0.5mm
    \centering
    \caption{Different GNN models on collaboration graph.}
    \label{ab_ogbn_table2}
    \resizebox{0.9\columnwidth}{!}{%0.9
        \begin{threeparttable}
        \begin{tabular}{lcc}
        \toprule
        \multirow{2}{*}{Methods} & non-overlapping &overlapping \\ 
                         & 20 clients   & 30 clients \\ 
        \midrule
        w/o Sheaf diffusion\tnote{a}    & 69.75 & 65.66 \\
        w/ GCN  \tnote{b}                 & 69.86 & 65.96 \\
        w/ GAT                   & 69.92 & 65.90 \\
        w/ Sheaf diffusion (Ours)       & 71.87 & 67.89 \\ 
        \bottomrule
        \end{tabular}
        \begin{tablenotes}
            \item[a] Remove the ``sheaf diffusion'' part in FedSheafHN.
            \item[b] Replace sheaf diffusion model with GCN.
        \end{tablenotes}
        \end{threeparttable}
        }       
\end{minipage}
\hfill
\begin{minipage}[!t]{0.4\textwidth} %0.48
\vspace{-3mm}
\begin{figure}[H]
    \centering
    \begin{subfigure}{\textwidth} 
        \centering
        \includegraphics[width=\textwidth]{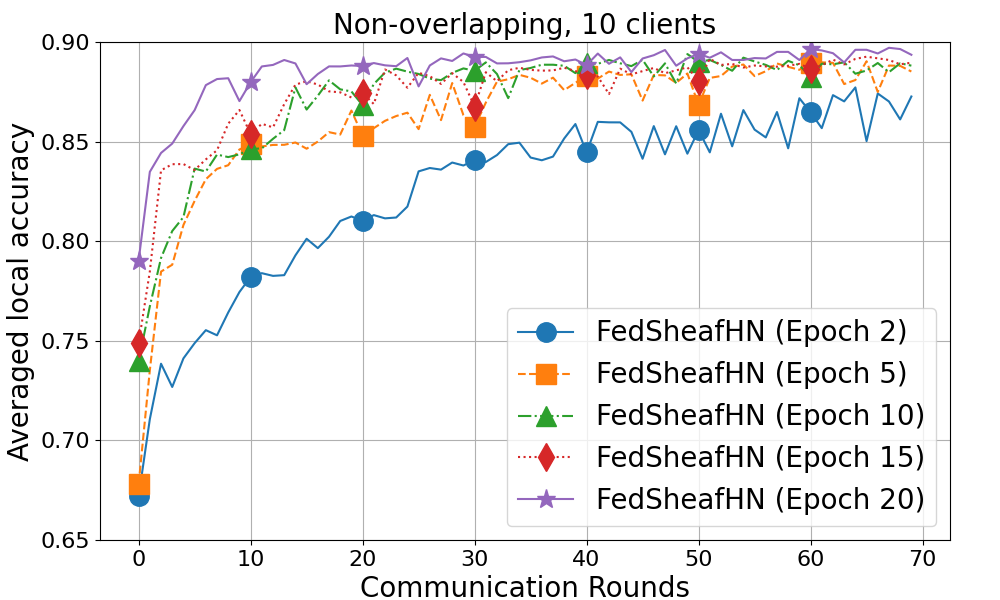}%0.7
        \vspace{-0.5mm}
    \end{subfigure}\\
    \begin{subfigure}{\textwidth} 
        \centering
        \includegraphics[width=\textwidth]{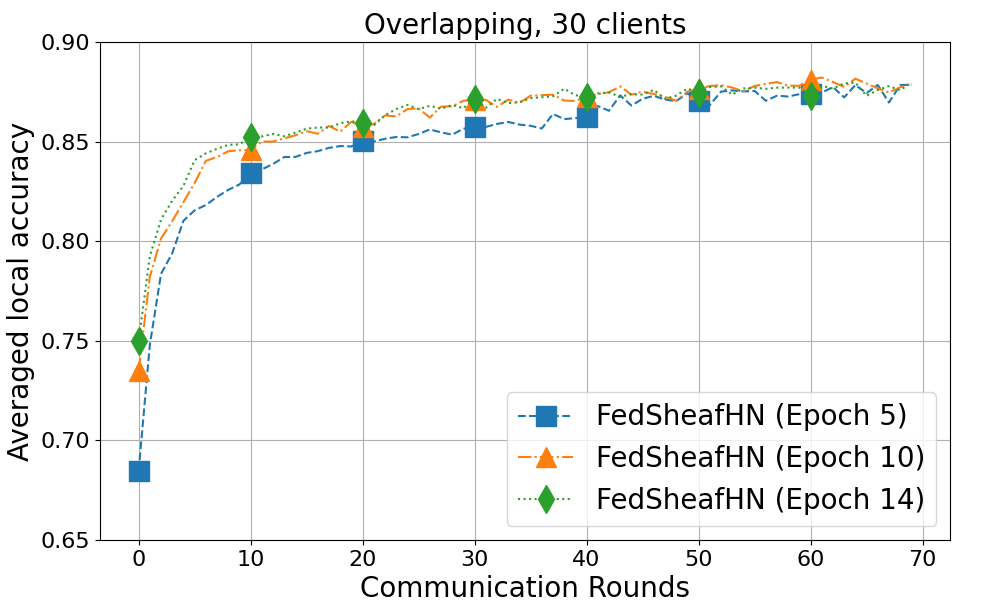}%0.7
        \vspace{-2mm}
    \end{subfigure}
    \vspace{-4mm}
    \caption{Results on Computer by varying local epochs in different scenarios.} %\red{[No need for ``Performance on  Computer'' in the plot titles.]}}
    \label{localep}
\end{figure}
\end{minipage}

\vspace{+0.5mm}
%%%%%%%%%%%%%%%%%%%%%%%%%%%%%%%%%%%%%%%
This emphasizes the importance of the collaboration graph with client graph-level embeddings, which provide refined client data-driven representations. These embeddings equip the hypernetwork to generate more precisely tailored model parameters, enhancing overall effectiveness. The interplay among these components underscores the synergistic nature of our approach, highlighting the crucial role of each in achieving superior FL performance.

We also compare the performance of different GNN models in improving the collaboration graph in \cref{ab_ogbn_table2}. Although the ground truth of the graph's edges is unknown, the sheaf diffusion process effectively explores underlying relationships through client graph-level embeddings, enabling it to aggregate information from other clients and enhance the embedding vectors more efficiently.
%######################################
\vspace{-3mm}
\paragraph{Varying Local Epochs.}  
\Cref{localep} demonstrate that increasing the number of local update steps can lead to local models diverging towards their respective subgraphs, thus underscoring the nuanced relationship between local training intensity and model convergence. Consequently, more local epochs do not always equate to better performance, and the optimal number of local epochs varies across different datasets. This highlights the need for a balanced approach to local training in FL.
%#########################################################################
\vspace{-3.5mm}
\section{Conclusion}
\vspace{-3.5mm}
In conclusion, our FedSheafHN framework, through its innovative use of collaboration graph and integration of sheaf diffusion with a hypernetwork featuring an attention layer, proves highly effective in personalized subgraph FL settings. This approach adeptly manages heterogeneity among clients, as evidenced by its superior performance compared to other baselines, fast convergence, and ability to generalize to new clients. Our ablation studies emphasize the significant contributions of each component, particularly the importance of client graph-level embedding in constructing the collaboration graph. However, the exploration of local update steps suggests a potential for model divergence, stressing the need to tailor local epochs for specific datasets. Overall, FedSheafHN presents a promising strategy for enhancing collaborative learning while accommodating the unique characteristics of each client within a federated network.
Limitations and impacts discussed in Appendix~\ref{app_discussion}.

\bibliographystyle{IEEEtran}
\bibliography{bibs/fedsheaf_ref,bibs/SIGNAL}
%%%%%%%%%%%%%%%%%%%%%%%%%%%%%%%%%%%%%%%%%%%%%%%%%%%%%%%%%%%%%%%%%%%%%%%%%%
\newpage
\appendix

\section{Experimental Setups}
\subsection{Datasets}
\label{app_datasets}
\begin{table*}[!ht]
\caption{Dataset statistics for the non-overlapping node scenario. We present data on the number of nodes, edges, classes, clustering coefficient, and heterogeneity for both the original graph and its split subgraphs under overlapping and disjoint node scenarios. ``Ori'' signifies the original largest connected components in the graph.}
\vspace{-5mm}
\label{nonoverlapping_dataset_statistics} 
\vskip 0.15in
\begin{center}
\footnotesize
\resizebox{\textwidth}{!}{
\begin{tabular}{lccccccccc}
\toprule
\multirow{2}{*}{} & \multicolumn{3}{c}{Cora}   & \multicolumn{3}{c}{Citeseer}   & \multicolumn{3}{c}{Pubmed}      \\ \cline{2-10} 
                         & Ori & 10 Clients & 20 Clients  & Ori & 10 Clients & 20 Clients  & Ori & 10 Clients & 20 Clients    \\ 
\midrule
\# Classes               & \multicolumn{3}{c}{7}       & \multicolumn{3}{c}{6}       & \multicolumn{3}{c}{3}               \\
\#  Nodes                & 2,485   &  249    &  124    &  2,120   &  212   & 106     &  19,717   &  1,972   &  986         \\
\# Edges                 & 10,138  &  891    &  422    &  7,358   &  675   & 326     &  88,648   &  7,671   & 3,607        \\
Clustering Coefficient   & 0.238   &  0.259  &  0.263  &  0.170   &  0.178 & 0.180   &  0.060    &  0.066   & 0.067        \\
Heterogeneity            & N/A     &  0.606  &  0.665  &  N/A     &  0.541 & 0.568   &  N/A      &  0.392   & 0.424        \\ 
\bottomrule
\multicolumn{10}{l}{}                                                                                                                         \\ 
\toprule
\multirow{2}{*}{} & \multicolumn{3}{c}{Amazon-Computer} & \multicolumn{3}{c}{Amazon-Photo}    & \multicolumn{3}{c}{ogbn-arxiv}        \\ \cline{2-10} 
                & Ori & 10 Clients & 20 Clients  & Ori & 10 Clients & 20 Clients  & Ori & 10 Clients & 20 Clients    \\ 
\midrule
\# Classes               & \multicolumn{3}{c}{10}       & \multicolumn{3}{c}{8}       & \multicolumn{3}{c}{40}               \\
\#  Nodes                &  13,381    &  1,338          &   669          & 7,487      &   749         &      374      &  169,343    &   16,934         & 8,467   \\
\# Edges                 &  491,556   &    36,136       &     15,632     & 238,086    &    19,322     &     8,547     &  2,315,598  &    182,226       &  86,755\\
Clustering Coefficient   &   0.351    &    0.398        &   0.418        &  0.410     &    0.457      &        0.477  &  0.226      &  0.259           &   0.269\\
Heterogeneity            &  N/A       &   0.612         &    0.647       &  N/A       &     0.681     &    0.751      & N/A         &    0.615         &  0.637        \\ 
\bottomrule
\end{tabular}
}
\end{center}
\end{table*}
\begin{table*}[!ht]
\caption{Dataset statistics for the overlapping node scenario.
We present data on the number of nodes, edges, classes, clustering coefficient, and heterogeneity for both the original graph and its split subgraphs.}
\label{overlapping_dataset_statistics}
\vspace{-4mm}
\begin{center}
\footnotesize
\resizebox{\textwidth}{!}{
\begin{tabular}{lccccccccc}
\toprule
\multirow{2}{*}{} & \multicolumn{3}{c}{Cora}   & \multicolumn{3}{c}{Citeseer}   & \multicolumn{3}{c}{Pubmed}  \\ \cline{2-10} 
                & Ori & 30 Clients & 50 Clients  & Ori & 30 Clients & 50 Clients  & Ori & 30 Clients & 50 Clients   \\ 
\midrule
\# Classes               & \multicolumn{3}{c}{7}       & \multicolumn{3}{c}{6}       & \multicolumn{3}{c}{3}               \\
\#  Nodes                & 2,485    &   207         &  124           &  2,120   &     177      &     106        
&   19,717   &    1,643        &     986          \\
\# Edges                 & 10,138   &   379         &    215         &  7,358   &   293       &     170        
&   88,648  &    3,374        &      1,903 \\
Clustering Coefficient   & 0.238    &  0.129          &   0.125           &  0.170   &    0.087      &   0.096          
&  0.060    &     0.034        &     0.035          \\
Heterogeneity            & N/A      &   0.567         &  0.613           &  N/A     &   0.494        &   0.547          
&  N/A   &     0.383        &    0.394   \\ 
\bottomrule
\multicolumn{10}{l}{}                                                                                                                         \\ 
\toprule
\multirow{2}{*}{} & \multicolumn{3}{c}{Amazon-Computer} & \multicolumn{3}{c}{Amazon-Photo}    & \multicolumn{3}{c}{ogbn-arxiv}        \\ \cline{2-10} 
                & Ori & 30 Clients & 50 Clients & Ori & 30 Clients & 50 Clients & Ori & 30 Clients & 50 Clients  \\ 
\midrule
\# Classes               & \multicolumn{3}{c}{10}       & \multicolumn{3}{c}{8}       & \multicolumn{3}{c}{40}               \\
\#  Nodes                &  13,381    &    1,115         &   669          & 7,487      &     624       &       374     &  169,343    &   14,112         &    8,467   \\
\# Edges                 &  491,556   &   16,684         &   8,969        & 238,086    &     8,735     &      4,840    &  2,315,598  &     83,770       &   44,712 \\
Clustering Coefficient   &   0.351    &    0.348         &   0.359        &  0.410     &    0.391      &      0.410    &  0.226      &  0.185           &   0.191      \\
Heterogeneity            &  N/A       &   0.577          &     0.614      &  N/A       &     0.696     &       0.684   & N/A         &    0.606         &   0.615       \\ 
\bottomrule
\end{tabular}
}
\end{center}
\vskip -0.1in
\end{table*}

In our work, we present thorough statistical analyses derived from six distinct benchmark datasets \cite{sen2008collective,mcauley2015image,shchur2018pitfalls,hu2020open}: Cora, CiteSeer, Pubmed, and ogbn-arxiv for citation graphs, as well as Computer and Photo for Amazon product graphs. These datasets serve as the foundation for our experimental investigations, covering both non-overlapping and overlapping and node scenarios, detailed in \cref{nonoverlapping_dataset_statistics,overlapping_dataset_statistics}. The table provides a comprehensive overview of key metrics for each subgraph, encompassing node and edge counts, class distribution, and clustering coefficients \cite{baek2023personalized}. Notably, the clustering coefficient, indicating the extent of node clustering within individual subgraphs, is calculated by first determining the coefficient \cite{watts1998collective} for all nodes and subsequently computing the average. In contrast, heterogeneity, reflecting dissimilarity among disjointed subgraphs, is measured by calculating the median Jenson-Shannon divergence of label distributions across all subgraph pairs. These statistics offer a detailed understanding of the structural intricacies and relationships within the benchmark datasets utilized in the experiments. In partitioning datasets, we randomly allocate 40\% of nodes for training, 30\% for validation, and another 30\% for testing across all datasets.

We delineate the procedure for partitioning the original graph into multiple subgraphs, tailored to the number of clients participating in FL. The METIS graph partitioning algorithm \cite{karypis1997metis} is employed to effectively segment the original graph into distinct subgraphs, offering control over the number of disjoint subgraphs through adjustable parameters. In scenarios with non-overlapping nodes, each client is assigned a unique disjoint subgraph directly derived from the METIS algorithm output. For example, setting the METIS parameter to 10 results in 10 distinct disjoint subgraphs, each allocated to an individual client.
Conversely, in scenarios involving overlapping nodes across subgraphs, the process begins by dividing the original graph into 6 and 10 disjoint subgraphs for configurations with 30 and 50 clients, respectively, utilizing the METIS algorithm. Subsequently, within each split subgraph, half of the nodes and their associated edges are randomly sampled, forming a subgraph assigned to a specific client. This iterative process is repeated 5 times, generating 5 distinct yet overlapped subgraphs for each split subgraph obtained from METIS. This meticulous approach ensures a varied yet controlled distribution of data across clients, accommodating both non-overlapping and overlapping node scenarios within the framework of FL.

%##########################################################
\vspace{-2mm}
\subsection{Comparison for Baselines}
\label{app_baselines}
\vspace{-1mm}
\paragraph{FedAvg.} 
The methodology, referenced as the FL baseline \cite{mcmahan2017communication}, involves a decentralized approach wherein each client independently updates a model and transmits it to a central server. The server then performs model aggregation, considering the varying numbers of training samples across clients, and subsequently communicates the aggregated model back to the individual clients. This FL baseline embodies a distributed learning paradigm, emphasizing local model updates and collaborative model aggregation at the central server. The essence lies in the iterative exchange between clients and the central server, where localized insights contribute to the collective improvement of the global model. Such a mechanism ensures a collaborative learning process while accommodating the inherent diversity in local datasets and training samples across participating clients. The FL baseline serves as a foundational framework for FL, embodying the principles of decentralized model training and collaborative knowledge sharing among distributed entities.
\vspace{-2mm}
\paragraph{FedPer.} 
As the embodiment of a PFL baseline, this method \cite{arivazhagan2019federated} introduces a unique approach: sharing only the base layers across participating clients while keeping personalized classification layers local. During collaborative learning, the neural network's foundational base layers are transmitted and synchronized globally, while the personalized classification layers, capturing individual nuances, remain decentralized within each client's local environment. FedPer's framework strikes a balance between global model synchronization and individualized feature representation. Sharing common base layers enables the model to benefit from collective insights, while retaining personalized classification layers ensures adaptation to client-specific patterns. Particularly valuable for preserving client privacy and accommodating diverse local data, FedPer navigates the nuanced landscape of collaborative learning by harmonizing global and local model components.
\vspace{-2mm}
\paragraph{FedSage+.}
This method \cite{zhang2021subgraph} is a subgraph FL baseline that uses a local graph generator to expand local subgraphs by generating new nodes. To train the generator, clients first receive node representations from other clients to enrich their own local subgraphs. They then compute gradients based on the differences between their local node features and the received representations. These gradients are sent back to the originating clients to help refine the graph generator, enhancing the ability to capture node relationships and improving the representation of each subgraph.
\vspace{-2mm}
\paragraph{pFedHN.}
This framework \cite{shamsian2021personalized} introduces an innovative solution to PFL, where the objective is to train distinct models for multiple clients, each with unique data distributions. pFedHN leverages a central hypernetwork that is tasked with generating a dedicated model for each client, enabling personalized model training while simultaneously addressing the disparities in data distribution across clients. The architecture allows for efficient parameter sharing, ensuring that each client receives a model that is both unique and optimally tuned to their specific data. A key advantage of using HNs in this setup is the significant reduction in communication costs, as the hypernetwork parameters themselves are not transmitted. This effectively decouples the communication overhead from the size of the model being trained. Additionally, the ability of hypernetworks to share information across clients enhances the generalizability of the framework, making pFedHN particularly adept at adapting to new clients with data distributions not seen during the initial training phase.
\vspace{-2mm}
\paragraph{pFedGraph.}
This method \cite{ye2023personalized} enhances PFL by addressing the challenges of data heterogeneity and malicious clients. The core innovation is the development of a collaboration graph that models the benefits of pairwise collaboration and distributes collaboration strengths among clients. The approach involves two main components: (1) the server infers the collaboration graph based on model similarity and dataset size, enabling detailed collaboration; and (2) clients optimize their local models with assistance from the server's aggregated model, promoting personalization. By learning a collaboration graph, pFedGraph adapts to different data heterogeneity levels and mitigates model poisoning attacks, guiding each client to collaborate primarily with similar and beneficial peers.
\vspace{-2mm}
\paragraph{FED-PUB.} 
This approach \cite{baek2023personalized} goes beyond estimating similarities between subgraphs solely based on their models' functional embeddings for community structure discovery. Additionally, it incorporates an adaptive masking mechanism for received weights from the server. This adaptive masking serves to filter out irrelevant weights originating from heterogeneous communities. By combining the estimation of similarities through functional embeddings with the adaptive filtering of received weights, FED-PUB enhances the accuracy of community structure discovery in a federated setting. This dual strategy ensures that the model not only captures similarities between subgraphs but also selectively focuses on relevant weights, thereby optimizing the identification of community structures within the diverse and distributed data across participating entities. 
\vspace{-2mm}
\paragraph{FedSheafHN.}
This is our Federated Learning with Neural Sheaf Diffusion and Hypernetwork (FedSheafHN) framework, which enhances heterogeneous graph learning models tailored to individual clients rather than pursuing a singular global model or sharing parameters extensively among clients. Within this model, we utilize graph-level embeddings from local GNNs to construct collaboration graph, employ sheaf diffusion to enhance this graph and use hypernetworks to dynamically generate distinct local graph models for clients with diverse and heterogeneous datasets. FedSheafHN adapts to the specific characteristics of each client's data, ensuring a personalized and effective learning process within the federated setting.  
%###########################################################
\section{Supplementary Experiments}
\label{app_exp}
\subsection{Malicious Clients}
We consider scenarios involving malicious clients aiming to disrupt the training process by sending arbitrary graph-level embedding to the server, with malicious client ratios set at 0.2, 0.4, 0.6 and 0.8. We consider two types of embedding for attack: same-value parameters generated by $x_{i}^{(ma)}=a\mathbf{1}_{d_{x_{i}}}$, where $\mathbf{1}_{d_{x_{i}}}\in \mathbb{R}^{d_{x_{i}}}$ is the vector of ones and $a\sim \mathcal{N}(0,\tau ^{2})$, and Gaussian noise parameters generated by $x_{i}^{(ma)}\sim \mathcal{N}(\mathbf{0} _{d_{x_{i}}},\tau^{2}I_{d_{x_{i}}})$ \cite{lin2022personalized}. The results are shown in \cref{attack_dis} and \cref{attack_over}.
\begin{table*}[h!]
\caption{Results of models under malicious client attacks on the ogbn-arxiv dataset with 20 clients in non-overlapping scenario.}
\label{attack_dis}
\small
\centering
\resizebox{\linewidth}{!}{
\begin{tabular}{c|c|c|c|c|c|c|c}
\toprule
\multicolumn{3}{c|}{\multirow{2}{*}{Type attack}} & \multicolumn{5}{c}{malicious client ratio}\\
\cline{4-8}
\multicolumn{3}{c|}{}& 0 & 0.2 & 0.4 & 0.6& 0.8\\
\hline
\multicolumn{1}{c|}{\multirow{4}{*}{Same value with different $\tau$}} & \multicolumn{1}{c|}{\multirow{2}{*}{$\tau=5$}}& \multicolumn{1}{c|}{\multirow{1}{*}{FedSheafHN}}&71.87 &71.56 &71.47 &71.44 &71.38 \\
\cline{3-8}
\multicolumn{1}{c|}{\multirow{4}{*}{}} & \multicolumn{1}{c|}{} &\multicolumn{1}{c|}{\multirow{1}{*}{FED-PUB}}&66.64 &60.43 &60.16 &60.38 &60.30 \\
\cline{2-8}
\multicolumn{1}{c|}{\multirow{4}{*}{}} & \multicolumn{1}{c|}{\multirow{2}{*}{$\tau=20$}}& \multicolumn{1}{c|}{\multirow{1}{*}{FedSheafHN}}&71.87 &71.16 &71.02 &70.93 &70.87 \\
\cline{3-8}
\multicolumn{1}{c|}{\multirow{4}{*}{}} & \multicolumn{1}{c|}{}& \multicolumn{1}{c|}{\multirow{1}{*}{FED-PUB}}&66.64 &60.49 &60.09 &60.14 &60.13 \\
\hline
\cline{3-8}
\multicolumn{1}{c|}{\multirow{4}{*}{Gaussian with different $\tau$}} & \multicolumn{1}{c|}{\multirow{2}{*}{$\tau=5$}}& \multicolumn{1}{c|}{\multirow{1}{*}{FedSheafHN}}& 71.87 &71.71 &71.56 &71.51 &71.41 \\
\cline{3-8}
\multicolumn{1}{c|}{\multirow{4}{*}{}} & \multicolumn{1}{c|}{}& \multicolumn{1}{c|}{\multirow{1}{*}{FED-PUB}}&66.64 &60.58 &61.25 &61.30 &60.94 \\
\cline{2-8}
\multicolumn{1}{c|}{\multirow{4}{*}{}} & \multicolumn{1}{c|}{\multirow{2}{*}{$\tau=20$}}& \multicolumn{1}{c|}{\multirow{1}{*}{FedSheafHN}}&71.87 &71.21 &71.15 &71.03 &70.94 \\
\cline{3-8}
\multicolumn{1}{c|}{\multirow{4}{*}{}} & \multicolumn{1}{c|}{}& \multicolumn{1}{c|}{\multirow{1}{*}{FED-PUB}}&66.64 &59.96 &60.35 &60.99 &61.13 \\
\bottomrule
\end{tabular}}
\vspace{-2mm}
\end{table*}
\begin{table*}[h!]
\caption{Results of models under malicious client attacks on the ogbn-arxiv dataset with 30 clients in overlapping scenario.}
\label{attack_over}
\small
\centering
\resizebox{\linewidth}{!}{
\begin{tabular}{c|c|c|c|c|c|c|c}
\toprule
\multicolumn{3}{c|}{\multirow{2}{*}{Type attack}} & \multicolumn{5}{c}{malicious client ratio}\\
\cline{4-8}
\multicolumn{3}{c|}{}& 0 & 0.2 & 0.4 & 0.6& 0.8\\
\hline
\multicolumn{1}{c|}{\multirow{4}{*}{Same value with different $\tau$}} & \multicolumn{1}{c|}{\multirow{2}{*}{$\tau=5$}}& \multicolumn{1}{c|}{\multirow{1}{*}{FedSheafHN}}&67.89 &67.83 &67.70 &67.61 &67.52 \\
\cline{3-8}
\multicolumn{1}{c|}{\multirow{4}{*}{}} & \multicolumn{1}{c|}{} &\multicolumn{1}{c|}{\multirow{1}{*}{FED-PUB}}&62.55 &59.08 &59.31 &59.30 &59.02 \\
\cline{2-8}
\multicolumn{1}{c|}{\multirow{4}{*}{}} & \multicolumn{1}{c|}{\multirow{2}{*}{$\tau=20$}}& \multicolumn{1}{c|}{\multirow{1}{*}{FedSheafHN}}&67.89 &67.43 &67.19 &67.10 &67.00 \\
\cline{3-8}
\multicolumn{1}{c|}{\multirow{4}{*}{}} & \multicolumn{1}{c|}{}& \multicolumn{1}{c|}{\multirow{1}{*}{FED-PUB}}&62.55 &59.72 &59.65 &58.92 &59.29 \\
\hline
\cline{3-8}
\multicolumn{1}{c|}{\multirow{4}{*}{Gaussian with different $\tau$}} & \multicolumn{1}{c|}{\multirow{2}{*}{$\tau=5$}}& \multicolumn{1}{c|}{\multirow{1}{*}{FedSheafHN}}&67.89  &67.73 &67.75 &67.62 &67.45 \\
\cline{3-8}
\multicolumn{1}{c|}{\multirow{4}{*}{}} & \multicolumn{1}{c|}{}& \multicolumn{1}{c|}{\multirow{1}{*}{FED-PUB}}&62.55 &59.16 &59.40 &59.69 &59.66 \\
\cline{2-8}
\multicolumn{1}{c|}{\multirow{4}{*}{}} & \multicolumn{1}{c|}{\multirow{2}{*}{$\tau=20$}}& \multicolumn{1}{c|}{\multirow{1}{*}{FedSheafHN}}&67.89 &67.23 &67.09 &66.97 &66.91 \\
\cline{3-8}
\multicolumn{1}{c|}{\multirow{4}{*}{}} & \multicolumn{1}{c|}{}& \multicolumn{1}{c|}{\multirow{1}{*}{FED-PUB}}&62.55 &59.61 &59.62 &59.78 &59.60 \\
\bottomrule
\end{tabular}}
\vspace{-2mm}
\end{table*}
%###########################################################
\section{Discussions}
\label{app_discussion}
\vspace{-1mm}
%%%%%%%%%%%%%%%%%%%%%%%%%%%%%%%%%%
\subsection{Limitations}
\label{limitations}
\vspace{-1mm}
Our personalized subgraph FL framework, FedSheafHN, exhibits broad applicability across heterogeneous graph types. 
While our experiments primarily focus on prevalent unipartite graphs, probing the effectiveness of FedSheafHN in alternative graph structures holds promising potential for future exploration. 
Unipartite graphs are chosen for their widespread usage in current applications. 
However, investigating FedSheafHN's performance on diverse graph types presents an exciting avenue for research, unraveling novel insights into its adaptability across a spectrum of heterogeneous configurations. 
This exploration aims to extend FedSheafHN's utility, contributing to a deeper understanding of personalized FL in varied graph environments.
%%%%%%%%%%%%%%%%%%%%%%%%%%%%%%%%%%
\vspace{-2mm}
\subsection{Potential Impacts}
\label{impact}
\vspace{-1mm}
The significance of the FL mechanism lies in its pivotal role in safeguarding user privacy. Despite being extensively explored in image and language domains, its attention in graph-related contexts remains notably limited. Our research adopts an approach to handle the unique challenges in graph FL, addressing issues like underlying client relationships through innovative strategies employing graph-level embedding, construct collaboration graph, applying sheaf diffusion and attention-driven hypernetworks. 

Our work bears the potential to positively impact society by contributing to various domains reliant on graph-structured data, such as social networks, recommendation systems, and patient networks. Specifically, we emphasize the crucial role of our FL method in social and recommendation networks. In current real-world applications, storing all user interactions on the server poses privacy and data leakage risks. Our framework offers a solution by avoiding the storage of user interaction data on the server and, instead, sharing locally trained GNN models derived from client local graphs.

Nevertheless, the transmission of model parameters from the client to the server introduces privacy concerns. While our work primarily focuses on assuming the transmissible of model parameters without compromising privacy, the broader research community should delve into evaluating the safety of these parameters. If found unsafe to share, additional measures such as employing differential privacy techniques may be necessary to enhance their security.

\end{document}

%% file: preamble.tex
\usepackage[T1]{fontenc}
\usepackage[utf8]{inputenc}
\usepackage{mathtools}

\usepackage{amssymb,mathrsfs}% Typical maths resource packages
\usepackage{amsthm}
\usepackage{bm}
\usepackage{scalerel}
\usepackage{nicefrac}
\usepackage{microtype} 
\usepackage[shortlabels]{enumitem}
\usepackage{graphicx}
\usepackage{epstopdf}
\DeclareGraphicsExtensions{.eps,.png,.jpg,.pdf}

\usepackage{url}
\usepackage{colortbl}
\usepackage{booktabs}
\usepackage{multirow}
\usepackage{colortbl,xcolor}
\usepackage{soul}
\usepackage{xparse,xstring}
\usepackage{calc}
\usepackage{etoolbox}

\makeatletter
\@ifpackageloaded{natbib}{
	\relax
}{
	\usepackage{cite}
}
\makeatother

\usepackage{array}
\newcolumntype{L}[1]{>{\raggedright\let\newline\\\arraybackslash\hspace{0pt}}m{#1}}
\newcolumntype{C}[1]{>{\centering\let\newline\\\arraybackslash\hspace{0pt}}m{#1}}
\newcolumntype{R}[1]{>{\raggedleft\let\newline\\\arraybackslash\hspace{0pt}}m{#1}}

\makeatletter
\let\MYcaption\@makecaption
\makeatother
\usepackage[font=footnotesize]{subcaption}
\makeatletter
\let\@makecaption\MYcaption
\makeatother

\usepackage{glossaries}
\makeatletter
\sfcode`\.1006

\let\oldgls\gls
\let\oldglspl\glspl

\newcommand\fussy@ifnextchar[3]{%
	\let\reserved@d=#1%
	\def\reserved@a{#2}%
	\def\reserved@b{#3}%
	\futurelet\@let@token\fussy@ifnch}
\def\fussy@ifnch{%
	\ifx\@let@token\reserved@d
		\let\reserved@c\reserved@a
	\else
		\let\reserved@c\reserved@b
	\fi
	\reserved@c}

\renewcommand{\gls}[1]{%
\oldgls{#1}\fussy@ifnextchar.{\@checkperiod}{\@}}
\renewcommand{\glspl}[1]{%
\oldglspl{#1}\fussy@ifnextchar.{\@checkperiod}{\@}}

\newcommand{\@checkperiod}[1]{%
	\ifnum\sfcode`\.=\spacefactor\else#1\fi
}

\robustify{\gls}
\robustify{\glspl}
\makeatother

\newacronym{wrt}{w.r.t.}{with respect to}
\newacronym{RHS}{R.H.S.}{right-hand side}
\newacronym{LHS}{L.H.S.}{left-hand side}
\newacronym{iid}{i.i.d.}{independent and identically distributed}
\newacronym{SOTA}{SOTA}{state-of-the-art}
\usepackage{float}

\ifx\notloadhyperref\undefined
	\ifx\loadbibentry\undefined
		\usepackage[hidelinks,hypertexnames=false]{hyperref}
	\else
		\usepackage{bibentry}
		\makeatletter\let\saved@bibitem\@bibitem\makeatother
		\usepackage[hidelinks,hypertexnames=false]{hyperref}
		\makeatletter\let\@bibitem\saved@bibitem\makeatother
	\fi
\else
	\ifx\loadbibentry\undefined
		\relax
	\else
		\usepackage{bibentry}
	\fi
\fi

\usepackage[capitalize]{cleveref}
\makeatletter
\def\cref@getref#1#2{%
  \expandafter\let\expandafter#2\csname r@#1@cref\endcsname%
  \expandafter\expandafter\expandafter\def%
    \expandafter\expandafter\expandafter#2%
    \expandafter\expandafter\expandafter{%
      \expandafter\@firstoffive#2}}% <-------- five
\def\cpageref@getref#1#2{%
  \expandafter\let\expandafter#2\csname r@#1@cref\endcsname%
  \expandafter\expandafter\expandafter\def%
    \expandafter\expandafter\expandafter#2%
    \expandafter\expandafter\expandafter{%
      \expandafter\@secondoffive#2}}% <----------- five
      
\AtBeginDocument{%
   \def\label@noarg#1{%
    \cref@old@label{#1}%
    \@bsphack%
    \edef\@tempa{{page}{\the\c@page}}%
    \setcounter{page}{1}%
    \edef\@tempb{\thepage}%
    \expandafter\setcounter\@tempa%
    \cref@constructprefix{page}{\cref@result}%
    \protected@write\@auxout{}%
      {\string\newlabel{#1@cref}{{\cref@currentlabel}%
      {[\@tempb][\arabic{page}][\cref@result]\thepage}{}{}{}}}% <----- five
    \@esphack}%
  \def\label@optarg[#1]#2{%
    \cref@old@label{#2}%
    \@bsphack%
    \edef\@tempa{{page}{\the\c@page}}%
    \setcounter{page}{1}%
    \edef\@tempb{\thepage}%
    \expandafter\setcounter\@tempa%
    \cref@constructprefix{page}{\cref@result}%
    \protected@edef\cref@currentlabel{%
      \expandafter\cref@override@label@type%
        \cref@currentlabel\@nil{#1}}%
    \protected@write\@auxout{}%
      {\string\newlabel{#2@cref}{{\cref@currentlabel}%
      {[\@tempb][\arabic{page}][\cref@result]\thepage}{}{}{}}}% <------- five
    \@esphack}%
    }  
\makeatother
\crefname{equation}{}{}
\Crefname{equation}{}{}
\crefname{claim}{claim}{claims}
\crefname{step}{step}{steps}
\crefname{line}{line}{lines}
\crefname{condition}{condition}{conditions}
\crefname{dmath}{}{}
\crefname{dseries}{}{}
\crefname{dgroup}{}{}
\crefname{page}{page}{pages}

\crefname{Problem}{Problem}{Problems}
\crefformat{Problem}{Problem~#2#1#3}
\crefrangeformat{Problem}{Problems~#3#1#4 to~#5#2#6}

\crefname{Theorem}{Theorem}{Theorems}
\crefname{Corollary}{Corollary}{Corollaries}
\crefname{Proposition}{Proposition}{Propositions}
\crefname{Lemma}{Lemma}{Lemmas}
\crefname{Definition}{Definition}{Definitions}
\crefname{Example}{Example}{Examples}
\crefname{Assumption}{Assumption}{Assumptions}
\crefname{Remark}{Remark}{Remarks}
\crefname{Rem}{Remark}{Remarks}
\crefname{remarks}{Remarks}{Remarks}
\crefname{Appendix}{Appendix}{Appendices}
\crefname{Supplement}{Supplement}{Supplements}
\crefname{Exercise}{Exercise}{Exercises}
\crefname{TheoremA}{Theorem}{Theorems}
\crefname{CorollaryA}{Corollary}{Corollaries}
\crefname{PropositionA}{Proposition}{Propositions}
\crefname{LemmaA}{Lemma}{Lemmas}
\crefname{DefinitionA}{Definition}{Definitions}
\crefname{ExampleA}{Example}{Examples}
\crefname{RemarkA}{Remark}{Remarks}
\crefname{AssumptionA}{Assumption}{Assumptions}
\crefname{ExerciseA}{Exercise}{Exercises}

\usepackage{crossreftools}
\ifx\notloadhyperref\undefined
\else
	\relax
\fi

\usepackage{algorithm}
\usepackage{algpseudocode}

\ifx\loadbreqn\undefined
	\relax
\else
	\usepackage{breqn}
\fi

\makeatletter
\def\cleartheorem#1{%
    \expandafter\let\csname#1\endcsname\relax
    \expandafter\let\csname c@#1\endcsname\relax
}
\def\clearthms#1{ \@for\tname:=#1\do{\cleartheorem\tname} }
\makeatother

\ifx\renewtheorem\undefined
	\ifx\useTheoremCounter\undefined
		\newtheorem{Theorem}{Theorem}
		\newtheorem{Corollary}{Corollary}
		\newtheorem{Proposition}{Proposition}
		
	\else

	\fi

\fi

\theoremstyle{remark}

\theoremstyle{plain}

\newcommand{\qednew}{\nobreak \ifvmode \relax \else
		\ifdim\lastskip<1.5em \hskip-\lastskip
			\hskip1.5em plus0em minus0.5em \fi \nobreak
		\vrule height0.75em width0.5em depth0.25em\fi}



\NewDocumentCommand{\movedownsub}{e{^_}}{%
	\IfNoValueTF{#1}{%
		\IfNoValueF{#2}{^{}}% neither ^ nor _, do nothing; if no ^ but _, add ^{}
	}{%
		^{#1}% add superscript if present
	}%
	\IfNoValueF{#2}{_{#2}}% add subscript if present
}

\let\latexchi\chi
\RenewDocumentCommand{\chi}{}{\latexchi\movedownsub}

\DeclareSymbolFont{bsfletters}{OT1}{cmss}{bx}{n}
\DeclareSymbolFont{ssfletters}{OT1}{cmss}{m}{n}
\DeclareMathSymbol{\bsfGamma}{0}{bsfletters}{'000}
\DeclareMathSymbol{\ssfGamma}{0}{ssfletters}{'000}
\DeclareMathSymbol{\bsfDelta}{0}{bsfletters}{'001}
\DeclareMathSymbol{\ssfDelta}{0}{ssfletters}{'001}
\DeclareMathSymbol{\bsfTheta}{0}{bsfletters}{'002}
\DeclareMathSymbol{\ssfTheta}{0}{ssfletters}{'002}
\DeclareMathSymbol{\bsfLambda}{0}{bsfletters}{'003}
\DeclareMathSymbol{\ssfLambda}{0}{ssfletters}{'003}
\DeclareMathSymbol{\bsfXi}{0}{bsfletters}{'004}
\DeclareMathSymbol{\ssfXi}{0}{ssfletters}{'004}
\DeclareMathSymbol{\bsfPi}{0}{bsfletters}{'005}
\DeclareMathSymbol{\ssfPi}{0}{ssfletters}{'005}
\DeclareMathSymbol{\bsfSigma}{0}{bsfletters}{'006}
\DeclareMathSymbol{\ssfSigma}{0}{ssfletters}{'006}
\DeclareMathSymbol{\bsfUpsilon}{0}{bsfletters}{'007}
\DeclareMathSymbol{\ssfUpsilon}{0}{ssfletters}{'007}
\DeclareMathSymbol{\bsfPhi}{0}{bsfletters}{'010}
\DeclareMathSymbol{\ssfPhi}{0}{ssfletters}{'010}
\DeclareMathSymbol{\bsfPsi}{0}{bsfletters}{'011}
\DeclareMathSymbol{\ssfPsi}{0}{ssfletters}{'011}
\DeclareMathSymbol{\bsfOmega}{0}{bsfletters}{'012}
\DeclareMathSymbol{\ssfOmega}{0}{ssfletters}{'012}

\makeatletter
\newcommand*\rel@kern[1]{\kern#1\dimexpr\macc@kerna}
\newcommand*\widebar[1]{%
  \begingroup
  \def\mathaccent##1##2{%
    \rel@kern{0.8}%
    \overline{\rel@kern{-0.8}\macc@nucleus\rel@kern{0.2}}%
    \rel@kern{-0.2}%
  }%
  \macc@depth\@ne
  \let\math@bgroup\@empty \let\math@egroup\macc@set@skewchar
  \mathsurround\z@ \frozen@everymath{\mathgroup\macc@group\relax}%
  \macc@set@skewchar\relax
  \let\mathaccentV\macc@nested@a
  \macc@nested@a\relax111{#1}%
  \endgroup
}
\makeatother

\DeclareMathOperator*{\argmin}{arg\,min}

\DeclareMathOperator{\var}{var}

\DeclareMathOperator{\cov}{cov}

\newcommand{\ifbcdot}[1]{\ifblank{#1}{\cdot}{#1}}

\DeclarePairedDelimiterX\abs[1]{\lvert}{\rvert}{\ifbcdot{#1}}
\DeclarePairedDelimiterX\parens[1]{(}{)}{\ifbcdot{#1}}
\DeclarePairedDelimiterX\brk[1]{[}{]}{\ifbcdot{#1}}
\DeclarePairedDelimiterX\braces[1]{\{}{\}}{\ifbcdot{#1}}
\DeclarePairedDelimiterX\angles[1]{\langle}{\rangle}{\ifblank{#1}{\cdot,\cdot}{#1}}
\DeclarePairedDelimiterX\ip[2]{\langle}{\rangle}{\ifbcdot{#1},\ifbcdot{#2}}
\DeclarePairedDelimiterX\norm[1]{\lVert}{\rVert}{\ifbcdot{#1}}
\DeclarePairedDelimiterX\ceil[1]{\lceil}{\rceil}{\ifbcdot{#1}}
\DeclarePairedDelimiterX\floor[1]{\lfloor}{\rfloor}{\ifbcdot{#1}}

\DeclareFontFamily{U}{matha}{\hyphenchar\font45}
\DeclareFontShape{U}{matha}{m}{n}{
      <5> <6> <7> <8> <9> <10> gen * matha
      <10.95> matha10 <12> <14.4> <17.28> <20.74> <24.88> matha12
      }{}
\DeclareSymbolFont{matha}{U}{matha}{m}{n}
\DeclareFontSubstitution{U}{matha}{m}{n}

\DeclareFontFamily{U}{mathx}{\hyphenchar\font45}
\DeclareFontShape{U}{mathx}{m}{n}{
      <5> <6> <7> <8> <9> <10>
      <10.95> <12> <14.4> <17.28> <20.74> <24.88>
      mathx10
      }{}
\DeclareSymbolFont{mathx}{U}{mathx}{m}{n}
\DeclareFontSubstitution{U}{mathx}{m}{n}

\DeclareMathDelimiter{\vvvert}{0}{matha}{"7E}{mathx}{"17}
\DeclarePairedDelimiterX\vertiii[1]{\vvvert}{\vvvert}{\ifbcdot{#1}}

\DeclarePairedDelimiterXPP\trace[1]{\operatorname{Tr}}{(}{)}{}{\ifbcdot{#1}} % column vector
\DeclarePairedDelimiterXPP\col[1]{\operatorname{col}}{\{}{\}}{}{\ifbcdot{#1}} % column vector
\DeclarePairedDelimiterXPP\row[1]{\operatorname{row}}{\{}{\}}{}{\ifbcdot{#1}} % row vector
\DeclarePairedDelimiterXPP\erf[1]{\operatorname{erf}}{(}{)}{}{\ifbcdot{#1}}
\DeclarePairedDelimiterXPP\erfc[1]{\operatorname{erfc}}{(}{)}{}{\ifbcdot{#1}}
\DeclarePairedDelimiterXPP\KLD[2]{D}{(}{)}{}{\ifbcdot{#1}\, \delimsize\|\, \ifbcdot{#2}} % KL divergence
\DeclarePairedDelimiterXPP\op[2]{\operatorname{#1}}{(}{)}{}{#2} % general operator

\DeclarePairedDelimiterXPP\indicate[1]{{\bf 1}}{\{}{\}}{}{\ifbcdot{#1}}

\NewDocumentCommand\ofrac{s m}{%
	\IfBooleanTF#1%
	{\dfrac{1}{#2}}%
	{\frac{1}{#2}}%
}
\NewDocumentCommand\ddfrac{s m m}{%
	\IfBooleanTF#1%
	{\dfrac{\mathrm{d} {#2}}{\mathrm{d} {#3}}}%
	{\frac{\mathrm{d} {#2}}{\mathrm{d} {#3}}}%
}
\NewDocumentCommand\ppfrac{s m m}{%
	\IfBooleanTF#1%
	{\dfrac{\partial {#2}}{\partial {#3}}}%
	{\frac{\partial {#2}}{\partial {#3}}}%
}

\newcommand{\setgiven}{:}
\providecommand\given{}
\DeclarePairedDelimiterX\Set[2]\{\}{%
	\if#1:%
		\renewcommand\given{\SetSymbol{:}}%
	\else%
		\renewcommand\given{\SetSymbol[\delimsize]{#1}}%
	\fi%
#2
}

\NewDocumentCommand\set{s O{\setgiven} m}{%
	\IfBooleanTF#1%
	{\Set*{#2}{#3}}%
	{\Set{#2}{#3}}%
}

\NewDocumentCommand{\evalat}{ s O{\big} m e{_^} }{%
\IfBooleanTF{#1}%
{\left. #3 \right|}{#3#2|}%
\IfValueT{#4}{_{#4}}%
\IfValueT{#5}{^{#5}}%
}

\providecommand\given{}
\DeclarePairedDelimiterXPP\cprob[1]{}(){}{
\renewcommand\given{\nonscript\,\delimsize\vert\allowbreak\nonscript\,\mathopen{}}%
#1%
}
\DeclarePairedDelimiterXPP\cexp[1]{}[]{}{
\renewcommand\given{\nonscript\,\delimsize\vert\allowbreak\nonscript\,\mathopen{}}%
#1%
}

\DeclareDocumentCommand \P { s e{_^} d() g } {%
	\mathbb{P}%
	\IfBooleanTF{#1}%
		{
			\IfValueT{#2}{_{#2}}%
			\IfValueT{#3}{^{#3}}%
			\IfValueTF{#5}{\cprob{#4 \given #5}}{\IfValueT{#4}{\cprob{#4}}}%
		}%
		{
			\IfValueT{#2}{_{#2}}%
			\IfValueT{#3}{^{#3}}%
			\IfValueTF{#5}{\cprob*{#4 \given #5}}{\IfValueT{#4}{\cprob*{#4}}}%
		}%
}

\DeclareDocumentCommand \E { s e{_^} o g } {%
	\mathbb{E}%
	\IfBooleanTF{#1}%
		{
			\IfValueT{#2}{_{#2}}%
			\IfValueT{#3}{^{#3}}%
			\IfValueTF{#5}{\cexp{#4 \given #5}}{\IfValueT{#4}{\cexp{#4}}}%
		}%
		{
			\IfValueT{#2}{_{#2}}%
			\IfValueT{#3}{^{#3}}%	
			\IfValueTF{#5}{\cexp*{#4 \given #5}}{\IfValueT{#4}{\cexp*{#4}}}%		
		}%
}

\DeclareDocumentCommand \Var { s e{_^} d() g } {%
	\var%
	\IfBooleanTF{#1}%
		{
			\IfValueT{#2}{_{#2}}%
			\IfValueT{#3}{^{#3}}%
			\IfValueTF{#5}{\cprob{#4 \given #5}}{\IfValueT{#4}{\cprob{#4}}}%
		}%
		{
			\IfValueT{#2}{_{#2}}%
			\IfValueT{#3}{^{#3}}%	
			\IfValueTF{#5}{\cprob*{#4 \given #5}}{\IfValueT{#4}{\cprob*{#4}}}%		
		}%
}

\DeclareDocumentCommand \Cov { s e{_^} d() g } {%
	\cov%
	\IfBooleanTF{#1}%
		{
			\IfValueT{#2}{_{#2}}%
			\IfValueT{#3}{^{#3}}%
			\IfValueTF{#5}{\cprob{#4 \given #5}}{\IfValueT{#4}{\cprob{#4}}}%
		}%
		{
			\IfValueT{#2}{_{#2}}%
			\IfValueT{#3}{^{#3}}%	
			\IfValueTF{#5}{\cprob*{#4 \given #5}}{\IfValueT{#4}{\cprob*{#4}}}%		
		}%
}

\ExplSyntaxOn
\NewDocumentCommand \dist {m o o} {%
\mathrm{#1}\left(%
	\IfValueT{#3}{%
		\tl_if_blank:nTF{ #3 }{\cdot\, \middle|\, }{#3\, \middle|\, }%
	}
	\IfValueT{#2}{#2}%
\right)%
}
\ExplSyntaxOff

\NewDocumentCommand {\cbrace} {t+ D[]{black} D(){\widthof{#5}} m m } {%
	\begingroup%
		\color{#2}
		\IfBooleanTF{#1}{%
			\overbrace{#4}^%
		}{
			\underbrace{#4}_%
		}%
		{\parbox[c]{#3}{\centering\footnotesize{#5}}}%
	\endgroup% 
}

\let\oldforall\forall
\renewcommand{\forall}{\oldforall \, }

\let\oldexist\exists
\renewcommand{\exists}{\oldexist \, }

\makeatletter
\newcommand{\udcloser}[1]{\underline{\smash{#1}}}

\newcommand{\rankcolor}[2]{%
	\expandafter\renewcommand\csname #1\endcsname[1]{%
		\ifblank{##1}{%
			{\color{#2} \textbf{#2}}%
		}{%
			\ifmmode
				\textcolor{#2}{\bm{##1}}%
			\else%
				{\color{#2} \textbf{##1}}%
			\fi	
		}%
	}
}

\rankcolor{first}{red}
\rankcolor{second}{blue}
\rankcolor{third}{cyan}
\makeatother

\graphicspath{{./Figures/}{./figures/}}
\DeclareDocumentCommand{\includeCroppedPdf}{ o O{./Figures/} m }{
	\IfFileExists{#2#3-crop.pdf}{}{%
		\immediate\write18{pdfcrop #2#3.pdf #2#3-crop.pdf}}%
	\includegraphics[#1]{#2#3-crop.pdf}
}

\makeatletter
\newcommand*{\addFileDependency}[1]{% argument=file name and extension
  \typeout{(#1)}
  \@addtofilelist{#1}
  \IfFileExists{#1}{}{\typeout{No file #1.}}
}
\makeatother

\definecolor{gray90}{gray}{0.9}
\def\colorlist{red,blue,brown,cyan,darkgray,gray,lightgray,green,lime,magenta,olive,orange,pink,purple,teal,violet,white,yellow}

\makeatletter
\def\startcomment{[}
\ifx\nohighlights\undefined
	\newcommand{\createcolor}[1]{%
			\expandafter\newcommand\csname #1\endcsname[1]{{\color{#1} ##1}}%
	}
	\newcommand{\msout}[1]{\text{\color{green} \st{\ensuremath{#1}}}}
	\newcommand{\del}[1]{{\color{green}\ifmmode \msout{#1}\else\st{#1}\fi}}
\else
	\newcommand{\createcolor}[1]{%
			\expandafter\newcommand\csname #1\endcsname[1]{%
				\noexpandarg%
				\StrChar{##1}{1}[\firstletter]%
				\if\firstletter\startcomment%
					\relax
				\else%
					##1
				\fi
			}%
	}
	\newcommand{\msout}[1]{}
	\newcommand{\del}[1]{}
\fi

\def\@tempa#1,{%
    \ifx\relax#1\relax\else
        \createcolor{#1}%
        \expandafter\@tempa
    \fi
}
\expandafter\@tempa\colorlist,\relax,
\makeatother

\newcommand{\hhide}[1]{}
\ifx\diagnoselabel\undefined
	\relax
\else
	\makeatletter
	\def\@testdef #1#2#3{%
		\def\reserved@a{#3}\expandafter \ifx \csname #1@#2\endcsname
			\reserved@a  \else
			\typeout{^^Jlabel #2 changed:^^J%
				\meaning\reserved@a^^J%
				\expandafter\meaning\csname #1@#2\endcsname^^J}%
			\@tempswatrue \fi}
	\makeatother
\fi

%% file: neurips_2024.bbl
\begin{thebibliography}{10}
\providecommand{\url}[1]{#1}
\csname url@samestyle\endcsname
\providecommand{\newblock}{\relax}
\providecommand{\bibinfo}[2]{#2}
\providecommand{\BIBentrySTDinterwordspacing}{\spaceskip=0pt\relax}
\providecommand{\BIBentryALTinterwordstretchfactor}{4}
\providecommand{\BIBentryALTinterwordspacing}{\spaceskip=\fontdimen2\font plus
\BIBentryALTinterwordstretchfactor\fontdimen3\font minus \fontdimen4\font\relax}
\providecommand{\BIBforeignlanguage}[2]{{%
\expandafter\ifx\csname l@#1\endcsname\relax
\typeout{** WARNING: IEEEtran.bst: No hyphenation pattern has been}%
\typeout{** loaded for the language `#1'. Using the pattern for}%
\typeout{** the default language instead.}%
\else
\language=\csname l@#1\endcsname
\fi
#2}}
\providecommand{\BIBdecl}{\relax}
\BIBdecl

\bibitem{hamilton2020graph}
W.~L. Hamilton, \emph{Graph representation learning}.\hskip 1em plus 0.5em minus 0.4em\relax Morgan \& Claypool Publishers, 2020.

\bibitem{zhang2021subgraph}
K.~Zhang, C.~Yang, X.~Li, L.~Sun, and S.~M. Yiu, ``Subgraph federated learning with missing neighbor generation,'' \emph{Advances in Neural Information Processing Systems}, vol.~34, pp. 6671--6682, 2021.

\bibitem{wu2021fedgnn}
C.~Wu, F.~Wu, Y.~Cao, Y.~Huang, and X.~Xie, ``{FedGNN}: Federated graph neural network for privacy-preserving recommendation,'' \emph{arXiv preprint arXiv:2102.04925}, 2021.

\bibitem{pillutla2022federated}
K.~Pillutla, K.~Malik, A.-R. Mohamed, M.~Rabbat, M.~Sanjabi, and L.~Xiao, ``Federated learning with partial model personalization,'' in \emph{International Conference on Machine Learning}.\hskip 1em plus 0.5em minus 0.4em\relax PMLR, 2022, pp. 17\,716--17\,758.

\bibitem{wu2020comprehensive}
Z.~Wu, S.~Pan, F.~Chen, G.~Long, C.~Zhang, and S.~Y. Philip, ``A comprehensive survey on graph neural networks,'' \emph{IEEE transactions on neural networks and learning systems}, vol.~32, no.~1, pp. 4--24, 2020.

\bibitem{bodnar2022neural}
C.~Bodnar, F.~Di~Giovanni, B.~Chamberlain, P.~Li{\`o}, and M.~Bronstein, ``Neural sheaf diffusion: A topological perspective on heterophily and oversmoothing in gnns,'' \emph{Advances in Neural Information Processing Systems}, vol.~35, pp. 18\,527--18\,541, 2022.

\bibitem{JiLeeMen23}
F.~Ji, S.~H. Lee, H.~Meng, K.~Zhao, J.~Yang, and W.~P. Tay, ``Leveraging label non-uniformity for node classification in graph neural networks,'' in \emph{Proc. International Conference on Machine Learning}, ser. Proc. Machine Learning Research, vol. 202.\hskip 1em plus 0.5em minus 0.4em\relax PMLR, Jul. 2023, pp. 14\,869--14\,885.

\bibitem{KanZhaSon23}
Q.~Kang, K.~Zhao, Y.~Song, S.~Wang, and W.~P. Tay, ``Node embedding from neural hamiltonian orbits in graph neural networks,'' in \emph{Proc. International Conference on Machine Learning}, ser. Proc. Machine Learning Research, vol. 202.\hskip 1em plus 0.5em minus 0.4em\relax PMLR, Jul. 2023, pp. 15\,786--15\,808.

\bibitem{smith2017federated}
V.~Smith, C.-K. Chiang, M.~Sanjabi, and A.~S. Talwalkar, ``Federated multi-task learning,'' \emph{Advances in neural information processing systems}, vol.~30, 2017.

\bibitem{ha2016hypernetworks}
D.~Ha, A.~Dai, and Q.~V. Le, ``Hypernetworks,'' \emph{arXiv preprint arXiv:1609.09106}, 2016.

\bibitem{shamsian2021personalized}
A.~Shamsian, A.~Navon, E.~Fetaya, and G.~Chechik, ``Personalized federated learning using hypernetworks,'' in \emph{International Conference on Machine Learning}.\hskip 1em plus 0.5em minus 0.4em\relax PMLR, 2021, pp. 9489--9502.

\bibitem{xu2023heterogeneous}
Z.~Xu, L.~Yang, and S.~Gu, ``Heterogeneous federated learning based on graph hypernetwork,'' in \emph{International Conference on Artificial Neural Networks}.\hskip 1em plus 0.5em minus 0.4em\relax Springer, 2023, pp. 464--476.

\bibitem{mcmahan2017communication}
B.~McMahan, E.~Moore, D.~Ramage, S.~Hampson, and B.~A. y~Arcas, ``Communication-efficient learning of deep networks from decentralized data,'' in \emph{Artificial intelligence and statistics}.\hskip 1em plus 0.5em minus 0.4em\relax PMLR, 2017, pp. 1273--1282.

\bibitem{li2020federated}
T.~Li, A.~K. Sahu, M.~Zaheer, M.~Sanjabi, A.~Talwalkar, and V.~Smith, ``Federated optimization in heterogeneous networks,'' \emph{Proceedings of Machine learning and systems}, vol.~2, pp. 429--450, 2020.

\bibitem{li2021model}
Q.~Li, B.~He, and D.~Song, ``Model-contrastive federated learning,'' in \emph{Proceedings of the IEEE/CVF conference on computer vision and pattern recognition}, 2021, pp. 10\,713--10\,722.

\bibitem{ye2023fedfm}
R.~Ye, Z.~Ni, C.~Xu, J.~Wang, S.~Chen, and Y.~C. Eldar, ``{FedFM}: Anchor-based feature matching for data heterogeneity in federated learning,'' \emph{IEEE Transactions on Signal Processing}, 2023.

\bibitem{chen2023spectral}
Z.~Chen, H.~H. Yang, T.~Quek, and K.~F.~E. Chong, ``Spectral co-distillation for personalized federated learning,'' in \emph{Advances in Neural Information Processing Systems}, 2023.

\bibitem{mohri2019agnostic}
M.~Mohri, G.~Sivek, and A.~T. Suresh, ``Agnostic federated learning,'' in \emph{International Conference on Machine Learning}.\hskip 1em plus 0.5em minus 0.4em\relax PMLR, 2019, pp. 4615--4625.

\bibitem{arivazhagan2019federated}
M.~G. Arivazhagan, V.~Aggarwal, A.~K. Singh, and S.~Choudhary, ``Federated learning with personalization layers,'' \emph{arXiv preprint arXiv:1912.00818}, 2019.

\bibitem{wang2019federated}
K.~Wang, R.~Mathews, C.~Kiddon, H.~Eichner, F.~Beaufays, and D.~Ramage, ``Federated evaluation of on-device personalization,'' \emph{arXiv preprint arXiv:1910.10252}, 2019.

\bibitem{schneider2021personalization}
J.~Schneider and M.~Vlachos, ``Personalization of deep learning,'' in \emph{Proceedings of International Data Science Conference}.\hskip 1em plus 0.5em minus 0.4em\relax Springer, 2021, pp. 89--96.

\bibitem{hanzely2020lower}
F.~Hanzely, S.~Hanzely, S.~Horv{\'a}th, and P.~Richt{\'a}rik, ``Lower bounds and optimal algorithms for personalized federated learning,'' \emph{Advances in Neural Information Processing Systems}, vol.~33, pp. 2304--2315, 2020.

\bibitem{hanzely2020federated}
F.~Hanzely and P.~Richt{\'a}rik, ``Federated learning of a mixture of global and local models,'' \emph{arXiv preprint arXiv:2002.05516}, 2020.

\bibitem{tan2022towards}
A.~Z. Tan, H.~Yu, L.~Cui, and Q.~Yang, ``Towards personalized federated learning,'' \emph{IEEE Transactions on Neural Networks and Learning Systems}, 2022.

\bibitem{deng2020adaptive}
Y.~Deng, M.~M. Kamani, and M.~Mahdavi, ``Adaptive personalized federated learning,'' \emph{arXiv preprint arXiv:2003.13461}, 2020.

\bibitem{mansour2020three}
Y.~Mansour, M.~Mohri, J.~Ro, and A.~T. Suresh, ``Three approaches for personalization with applications to federated learning,'' \emph{arXiv preprint arXiv:2002.10619}, 2020.

\bibitem{ma2022layer}
X.~Ma, J.~Zhang, S.~Guo, and W.~Xu, ``Layer-wised model aggregation for personalized federated learning,'' in \emph{Proceedings of the IEEE/CVF conference on computer vision and pattern recognition}, 2022, pp. 10\,092--10\,101.

\bibitem{jiang2019improving}
Y.~Jiang, J.~Kone{\v{c}}n{\`y}, K.~Rush, and S.~Kannan, ``Improving federated learning personalization via model agnostic meta learning,'' \emph{arXiv preprint arXiv:1909.12488}, 2019.

\bibitem{fallah2020personalized}
A.~Fallah, A.~Mokhtari, and A.~Ozdaglar, ``Personalized federated learning with theoretical guarantees: A model-agnostic meta-learning approach,'' \emph{Advances in Neural Information Processing Systems}, vol.~33, pp. 3557--3568, 2020.

\bibitem{lee2024fedl2p}
R.~Lee, M.~Kim, D.~Li, X.~Qiu, T.~Hospedales, F.~Husz{\'a}r, and N.~Lane, ``Fedl2p: Federated learning to personalize,'' \emph{Advances in Neural Information Processing Systems}, vol.~36, 2024.

\bibitem{bui2019federated}
D.~Bui, K.~Malik, J.~Goetz, H.~Liu, S.~Moon, A.~Kumar, and K.~G. Shin, ``Federated user representation learning,'' \emph{arXiv preprint arXiv:1909.12535}, 2019.

\bibitem{collins2021exploiting}
L.~Collins, H.~Hassani, A.~Mokhtari, and S.~Shakkottai, ``Exploiting shared representations for personalized federated learning,'' in \emph{International conference on machine learning}.\hskip 1em plus 0.5em minus 0.4em\relax PMLR, 2021, pp. 2089--2099.

\bibitem{agarwal2021skellam}
N.~Agarwal, P.~Kairouz, and Z.~Liu, ``The skellam mechanism for differentially private federated learning,'' \emph{Advances in Neural Information Processing Systems}, vol.~34, pp. 5052--5064, 2021.

\bibitem{noble2022differentially}
M.~Noble, A.~Bellet, and A.~Dieuleveut, ``Differentially private federated learning on heterogeneous data,'' in \emph{International Conference on Artificial Intelligence and Statistics}.\hskip 1em plus 0.5em minus 0.4em\relax PMLR, 2022, pp. 10\,110--10\,145.

\bibitem{li2024clients}
Y.~Li, T.~Wang, C.~Chen, J.~Lou, B.~Chen, L.~Yang, and Z.~Zheng, ``Clients collaborate: Flexible differentially private federated learning with guaranteed improvement of utility-privacy trade-off,'' \emph{arXiv preprint arXiv:2402.07002}, 2024.

\bibitem{ghosh2020efficient}
A.~Ghosh, J.~Chung, D.~Yin, and K.~Ramchandran, ``An efficient framework for clustered federated learning,'' \emph{Advances in Neural Information Processing Systems}, vol.~33, pp. 19\,586--19\,597, 2020.

\bibitem{huang2021personalized}
Y.~Huang, L.~Chu, Z.~Zhou, L.~Wang, J.~Liu, J.~Pei, and Y.~Zhang, ``Personalized cross-silo federated learning on non-iid data,'' in \emph{Proceedings of the AAAI conference on artificial intelligence}, vol.~35, no.~9, 2021, pp. 7865--7873.

\bibitem{zhang2020personalized}
M.~Zhang, K.~Sapra, S.~Fidler, S.~Yeung, and J.~M. Alvarez, ``Personalized federated learning with first order model optimization,'' \emph{arXiv preprint arXiv:2012.08565}, 2020.

\bibitem{zhang2021parameterized}
J.~Zhang, S.~Guo, X.~Ma, H.~Wang, W.~Xu, and F.~Wu, ``Parameterized knowledge transfer for personalized federated learning,'' \emph{Advances in Neural Information Processing Systems}, vol.~34, pp. 10\,092--10\,104, 2021.

\bibitem{he2021fedgraphnn}
C.~He, K.~Balasubramanian, E.~Ceyani, C.~Yang, H.~Xie, L.~Sun, L.~He, L.~Yang, P.~S. Yu, Y.~Rong \emph{et~al.}, ``Fedgraphnn: A federated learning system and benchmark for graph neural networks,'' \emph{arXiv preprint arXiv:2104.07145}, 2021.

\bibitem{wang2022federatedscope}
Z.~Wang, W.~Kuang, Y.~Xie, L.~Yao, Y.~Li, B.~Ding, and J.~Zhou, ``Federatedscope-gnn: Towards a unified, comprehensive and efficient package for federated graph learning,'' in \emph{Proceedings of the 28th ACM SIGKDD Conference on Knowledge Discovery and Data Mining}, 2022, pp. 4110--4120.

\bibitem{xie2021federated}
H.~Xie, J.~Ma, L.~Xiong, and C.~Yang, ``Federated graph classification over non-iid graphs,'' \emph{Advances in Neural Information Processing Systems}, vol.~34, pp. 18\,839--18\,852, 2021.

\bibitem{he2021spreadgnn}
C.~He, E.~Ceyani, K.~Balasubramanian, M.~Annavaram, and S.~Avestimehr, ``Spreadgnn: Serverless multi-task federated learning for graph neural networks,'' \emph{arXiv preprint arXiv:2106.02743}, 2021.

\bibitem{tan2023federated}
Y.~Tan, Y.~Liu, G.~Long, J.~Jiang, Q.~Lu, and C.~Zhang, ``Federated learning on non-iid graphs via structural knowledge sharing,'' in \emph{Proceedings of the AAAI conference on artificial intelligence}, 2023, pp. 9953--9961.

\bibitem{yao2022fedgcn}
Y.~Yao, W.~Jin, S.~Ravi, and C.~Joe-Wong, ``Fedgcn: Convergence and communication tradeoffs in federated training of graph convolutional networks,'' \emph{arXiv preprint arXiv:2201.12433}, 2022.

\bibitem{baek2023personalized}
J.~Baek, W.~Jeong, J.~Jin, J.~Yoon, and S.~J. Hwang, ``Personalized subgraph federated learning,'' in \emph{International Conference on Machine Learning}.\hskip 1em plus 0.5em minus 0.4em\relax PMLR, 2023, pp. 1396--1415.

\bibitem{klein2015dynamic}
B.~Klein, L.~Wolf, and Y.~Afek, ``A dynamic convolutional layer for short range weather prediction,'' in \emph{Proceedings of the IEEE Conference on Computer Vision and Pattern Recognition}, 2015, pp. 4840--4848.

\bibitem{klocek2019hypernetwork}
S.~Klocek, {\L}.~Maziarka, M.~Wo{\l}czyk, J.~Tabor, J.~Nowak, and M.~{\'S}mieja, ``Hypernetwork functional image representation,'' in \emph{International Conference on Artificial Neural Networks}.\hskip 1em plus 0.5em minus 0.4em\relax Springer, 2019, pp. 496--510.

\bibitem{navon2020learning}
A.~Navon, A.~Shamsian, G.~Chechik, and E.~Fetaya, ``Learning the pareto front with hypernetworks,'' \emph{arXiv preprint arXiv:2010.04104}, 2020.

\bibitem{brock2017smash}
A.~Brock, T.~Lim, J.~M. Ritchie, and N.~Weston, ``{SMASH}: one-shot model architecture search through hypernetworks,'' \emph{arXiv preprint arXiv:1708.05344}, 2017.

\bibitem{zhang2018graph}
C.~Zhang, M.~Ren, and R.~Urtasun, ``Graph hypernetworks for neural architecture search,'' \emph{arXiv preprint arXiv:1810.05749}, 2018.

\bibitem{curry2014sheaves}
J.~M. Curry, \emph{Sheaves, cosheaves and applications}.\hskip 1em plus 0.5em minus 0.4em\relax University of Pennsylvania, 2014.

\bibitem{hansen2020sheaf}
J.~Hansen and T.~Gebhart, ``Sheaf neural networks,'' \emph{arXiv preprint arXiv:2012.06333}, 2020.

\bibitem{sen2008collective}
P.~Sen, G.~Namata, M.~Bilgic, L.~Getoor, B.~Galligher, and T.~Eliassi-Rad, ``Collective classification in network data,'' \emph{AI magazine}, vol.~29, no.~3, pp. 93--93, 2008.

\bibitem{hu2020open}
W.~Hu, M.~Fey, M.~Zitnik, Y.~Dong, H.~Ren, B.~Liu, M.~Catasta, and J.~Leskovec, ``Open graph benchmark: Datasets for machine learning on graphs,'' \emph{Advances in neural information processing systems}, vol.~33, pp. 22\,118--22\,133, 2020.

\bibitem{mcauley2015image}
J.~McAuley, C.~Targett, Q.~Shi, and A.~Van Den~Hengel, ``Image-based recommendations on styles and substitutes,'' in \emph{Proceedings of the 38th international ACM SIGIR conference on research and development in information retrieval}, 2015, pp. 43--52.

\bibitem{shchur2018pitfalls}
O.~Shchur, M.~Mumme, A.~Bojchevski, and S.~G{\"u}nnemann, ``Pitfalls of graph neural network evaluation,'' \emph{arXiv preprint arXiv:1811.05868}, 2018.

\bibitem{karypis1997metis}
G.~Karypis, ``Metis: Unstructured graph partitioning and sparse matrix ordering system,'' \emph{Technical report}, 1997.

\bibitem{ye2023personalized}
R.~Ye, Z.~Ni, F.~Wu, S.~Chen, and Y.~Wang, ``Personalized federated learning with inferred collaboration graphs,'' in \emph{International Conference on Machine Learning}.\hskip 1em plus 0.5em minus 0.4em\relax PMLR, 2023, pp. 39\,801--39\,817.

\bibitem{kingma2014adam}
D.~P. Kingma and J.~Ba, ``Adam: A method for stochastic optimization,'' \emph{arXiv preprint arXiv:1412.6980}, 2014.

\bibitem{watts1998collective}
D.~J. Watts and S.~H. Strogatz, ``Collective dynamics of ‘small-world’networks,'' \emph{nature}, vol. 393, no. 6684, pp. 440--442, 1998.

\bibitem{lin2022personalized}
S.~Lin, Y.~Han, X.~Li, and Z.~Zhang, ``Personalized federated learning towards communication efficiency, robustness and fairness,'' \emph{Advances in Neural Information Processing Systems}, vol.~35, pp. 30\,471--30\,485, 2022.

\end{thebibliography}
